\documentclass[10pt, conference]{IEEEtran}
\IEEEoverridecommandlockouts

\usepackage{cite}
\usepackage{amsthm, amsmath,amssymb,amsfonts}
\usepackage{graphicx}
\usepackage{textcomp}
\usepackage{xcolor}
\usepackage{hyperref}       
\usepackage{url}            
\usepackage{booktabs}       
\usepackage{amsfonts}       
\usepackage{nicefrac}       
\usepackage{microtype}      
\usepackage{wrapfig}
\usepackage{subfigure}
\usepackage{multirow}
\usepackage{algorithm}
\usepackage{algorithmicx}
\usepackage{algpseudocode}
\usepackage{caption}
\usepackage{enumitem}
\usepackage{array}
\usepackage{makecell}
\newtheorem{Definition}{Definition}
\newtheorem{Theorem}{Theorem}
\newtheorem{Lemma}{Lemma}
\newtheorem{Assumption}{Assumption}

\def\BibTeX{{\rm B\kern-.05em{\sc i\kern-.025em b}\kern-.08em
    T\kern-.1667em\lower.7ex\hbox{E}\kern-.125emX}}

\begin{document}

\title{Faster, Smaller, and Smarter: Task-Aware Expert Merging for Online MoE Inference
\thanks{
Corresponding author: Xutong Liu and Ruiting Zhou.

The work of Ruiting Zhou was supported in part by National Key Research and Development Program of China under Grant No. 2024YFB2907100, Shenzhen Science and Technology Program under Grant No. KJZD20240903100814018, National Natural Science Foundation of China under Grant No. 62232004.
The work of Xiangxiang Dai was supported by the National Natural Science Foundation of China (625B2163). The work of John C.S. Lui was supported in part by the RGC GRF-14202923.}
}

\author{\IEEEauthorblockN{
		Ziyi Han\IEEEauthorrefmark{1}, 
		Xutong Liu\IEEEauthorrefmark{2},
		Ruiting Zhou\IEEEauthorrefmark{3}, 
		Xiangxiang Dai\IEEEauthorrefmark{1},  
		John C.S. Lui\IEEEauthorrefmark{1}, 
	}
	\IEEEauthorblockA{ \IEEEauthorrefmark{1}The Chinese University of Hong Kong, 
		\IEEEauthorrefmark{2}University of Washington,
		\IEEEauthorrefmark{3}Southeast University\\
		Email: \{zyhan24, xxdai23, cslui\}@cse.cuhk.edu.hk, xutongl@uw.edu, ruitingzhou@seu.edu.cn.
	}
	
}

\maketitle

\begin{abstract}
Sparse Mixture of Experts (SMoE) has become a preferred architecture for scaling Transformer capacity without increasing computational cost, as it activates only a small subset of experts for each input. 
However, deploying such an approach for \textit{online inference} remains challenging due to the large size of a full SMoE model and the complexity of expert routing, especially in resource-constrained edge networks. Moreover, during the online inference, task information is often unavailable, making the task-level routing error-prone.
In this work, we propose a novel tree-structured adaptive neural bandit router, \texttt{Tanbr}, to enable efficient and reliable online MoE inference. 
Instead of relying on explicit task tags, \texttt{Tanbr} estimates the task distribution over time from historical data and uses it to guide task-aware expert merging within a given pre-trained MoE.
To handle the large continuous space of merging weights, \texttt{Tanbr} employs a binary tree to progressively partition the space and generate finer candidate weights. It then applies a neural bandit to learn the non-linear mapping from merging weights to model performance and decides optimal expert merging.
We prove that \texttt{Tanbr} achieves a sublinear regret bound of {\small $\mathcal{O}(\sqrt{T} \log(T))$} over {\small $T$} rounds, 
despite operating over a continuous decision space, 
matching regret bounds compared to existing methods.
Extensive experiments show that \texttt{Tanbr} reduces inference latency by at least {\small $45\%$} and memory usage by up to {\small $25\%$}, while maintaining a high accuracy compared to many state-of-the-art methods.
\end{abstract}


\section{Introduction}\label{sec:intro}
Modern deep learning models, particularly large language models (LLMs), have shown remarkable performance in a variety of tasks \cite{devlin2019bert, raffel2020exploring, dai2025multi}. However, this advancement has come at the cost of increasing computational demands as model capacity has grown exponentially by approximately {\small $10\times$} each year \cite{huang2024toward}.
To address these scalability challenges, Sparse Mixture of Experts (SMoE) has emerged as a promising architecture. By routing inputs to a small subset of experts, SMoE models efficiently reduce training costs while improving accuracy for tasks such as language modeling \cite{fedus2022switch}, machine translation \cite{costa2022no}, and image classification \cite{huang2023experts}.
Beyond training efficiency, the inference efficiency of MoE has received increasing attention due to the widespread deployment of large pre-trained models~\cite{liu2025optimizing, jin2025moe}. 
To this end, task-level routing has emerged as a promising strategy. It assigns entire tasks to specific experts based on task tags, improving both training and inference efficiency \cite{kudugunta2021beyond}. Compared to token-level routing, this approach reduces the frequency of router activations and expert switching, thereby lowering inference latency. 
As shown in Tab.~\ref{tab:motivation} (line 1 {\em vs.} 2 and 3 {\em vs.} 4), it achieves an efficiency gain of about {\small $35\%$}.

\vspace{-3mm}
\begin{table}[htb]
	\centering
	\caption{Comparison of various routing methods with T5-based MoE. All settings are the same as described in Sec \ref{sec:experiment}.
	}
	\label{tab:motivation}
	\vspace{-2mm}
	{
		\footnotesize
		\begin{tabular}{|c|c|c|}
			\hline
			\multirow{2}{*}{\textbf{Methods}}  & \textbf{Model Size } & \textbf{Inference Latency (s)} \\
			& full / loaded & active 1/2/3 experts\\
			\hline
			\makecell[l]{1. Token-SMoE \cite{fedus2022switch} }& 1.0B / 1.0B   & 1.81 / 2.42 / 3.05 \\
			\hline
			\makecell[l]{2. Task-SMoE \cite{kudugunta2021beyond}}  & 1.0B / 1.0B   & 1.16 / 1.78 / 2.41 \\
			\hline
			\makecell[l]{3. Token-Merging \cite{muqeeth2023soft}} & 1.0B / 1.0B   & 193.88 \\
			\hline
			\makecell[l]{4. Task-Merging \cite{he2023merging}} & 1.0B / 1.0B   & 3.01 \\
			\hline
			\makecell[l]{5. \texttt{Tanbr} (Ours)} & 1.0B / 220M  & 0.61 \\
			\hline
	\end{tabular}}
	\vspace{-3mm}
\end{table}

However, task-routing MoE under the \textit{online inference} setting faces several challenges that limit its adoption.
\textit{First}, varying input task distributions may lead to workload imbalance, where certain experts are activated more often than others. 
Replicating popular experts can alleviate this issue, but it introduces additional memory and synchronization overhead.
\textit{Second}, while selecting multiple experts can improve performance \cite{he2023merging}, it also significantly increases computational cost, as reflected in the increased inference latency in Tab. \ref{tab:motivation} (lines 1-2).
\textit{Third}, to avoid the overhead of loading and switching experts, the full MoE model is typically preloaded into memory. This leads to high memory demand and poses challenges for deployment in resource-limited environments, such as edge networks.
\textit{Finally}, task-routing MoE relies on the availability of task information ({\em e.g.}, tags) to identify the task type and route inputs to suitable experts.
However,  such information is often unavailable during online inference, making task-based routing difficult to apply.

To address the first two challenges, a promising approach is to merge experts by computing a weighted average of their parameters, so that the system runs a single merged expert instead of activating a subset of experts \cite{muqeeth2023soft,zhong2024lory}. This strategy incurs only the cost of running a single merged expert while leveraging the knowledge of multiple experts and prevents workload imbalance. However, existing expert merging methods either operate at the token level \cite{muqeeth2023soft}, incurring significant merging overhead (as shown in line 3 of Tab.~\ref{tab:motivation}), or rely on task tags \cite{he2023merging} or labeled datasets \cite{li2024merge}, limiting their applicability in the online setting. 
Additionally, some methods bypass explicit task tags by training classifiers to infer task types from input features \cite{li2025theory}. However, training and deploying such classifiers may incur substantial computational cost, limiting efficiency gains in resource-constrained online settings. 
These challenges lead us to ask the following research question: 

\textit{How to design a router that can effectively guide task-level expert merging within a given pre-trained MoE for online inference, even when explicit task tags are unavailable?}

While explicit task tags are often unavailable during inference, it is still possible to learn merging strategies by observing system feedback over time. This motivates the adoption of \textit{an online learning approach}.
In this paper, we present \texttt{Tanbr}, a novel neural bandit framework with continuous decision spaces of weights to guide expert merging for efficient and reliable online MoE inference.
To the best of our knowledge, this is the \textit{first work} to provide interoperable online MoE inference with formal theoretical guarantees.
Rather than relying on specific task tags, \texttt{Tanbr} captures task characteristics by using the task distribution within each time slot, which can be easily obtained in online environments using lightweight tools such as Count-Min Sketch \cite{cormode2005improved}.
Using the task distribution, \texttt{Tanbr} performs expert merging once per time slot, enabling a single merged model to serve all tasks within that slot. 
This reduces the overhead of frequent merging and lowers memory usage.
We adopt the Multi-Armed Bandit (MAB) theory to design the router for online learning. Unlike standard MAB settings, the router must handle a \textit{continuous and multi-dimensional} decision space of merging weights and a \textit{non-linear} mapping from merging weights to model performance.
To overcome this, we propose a tree-structured partitioning method that incrementally refines the continuous decision space while controlling search complexity.
We further develop a neural bandit that models the multi-dimensional reward and applies the Upper Confidence Bound (UCB) strategy to decide the optimal merging weight.
Our contributions are:
\begin{itemize}[noitemsep, topsep=0pt, left=.3em]
\item \textbf{Novel Model.} We formulate router design as a contextual bandit problem over a continuous space of merging weights. The context is the task distribution at each time slot, and the reward function can capture both linear and non-linear effects of the merging weights on model performance.
\item \textbf{Tree-structured Adaptive Neural Bandit Router}. To efficiently explore the continuous decision space, \texttt{Tanbr} uses a hierarchical partition tree to generate candidate merging weights. When a node is explored enough times, it expands to provide finer candidate merging weights.
\texttt{Tanbr} then utilizes a neural bandit that maintains a neural network to estimate the rewards of these candidate merging weights and selects the optimal one. Through rigorous theoretical analysis, we formally prove that \texttt{Tanbr} achieves a sublinear regret bound of {\small $\mathcal{O}(\tilde{d} \sqrt{T} \log(T))$}, where {\small $\tilde{d}$} denotes the effective dimension of the neural tangent kernel (NTK) matrix for the neural bandit network and {\small $T$} is the total number of system time slots.
Unlike previous neural bandit methods limited to discrete decisions, \texttt{Tanbr} extends to a large continuous decision space and still achieves comparable regret bounds. 
\item \textbf{Evaluation Performance}.  Extensive experiments 
show that: i) \texttt{Tanbr} exhibits robust offline/online learning abilities, converging rapidly and performing well on metrics like loss and accuracy; ii)  \texttt{Tanbr} reduces memory usage by up to {\small $25\%$} and improves inference efficiency by at least {\small $45\%$} while maintaining the same or even better accuracy compared to seven state-of-the-art baselines.
\end{itemize}

\section{Related Work}\label{sec:related}

\textbf{\textit{Router Design in MoE}.}
The router is a key component in all MoE architectures, controlling both the selection of expert computations and the combination of their outputs. To reduce computational and memory costs, two main types of router designs have been explored: SMoE and soft expert merging.
SMoE was first introduced by Shazeer {\em et al.} \cite{shazeer2017outrageously}. It improves model capacity and computational efficiency by activating only a small subset of experts for each input. Since then, many works have aimed to enhance the routing function to improve scalability, stability, and performance \cite{fedus2022switch, liu2023janus, shi2023pipe, wang2025spmoe}. 
For fine-tuning and inference efficiency,
Task-MoE \cite{kudugunta2021beyond} reduces inference costs by assigning entire tasks to a fixed subset of experts. 
Liu {\em et al.}  \cite{liu2025optimizing} propose a Bayesian optimization framework to efficiently learn expert selection strategies for distributed MoE inference in serverless environments.
To address gradient computation challenges during training, Muqeeth {\em et al.} \cite{muqeeth2023soft} propose soft expert merging methods, where the router outputs weights to merge expert parameters. The merged parameters are then used for forward computation and backpropagation to update the expert models. Extending this idea, Lory \cite{zhong2024lory} incorporates input similarity at the sequence level to guide expert merging. To improve inference efficiency, He {\em et al.} \cite{he2023merging} merge experts into a unified model based on task-specific tags. Li {\em et al.} \cite{li2024merge} design expert grouping and merging strategies that require substantial training data to assess expert similarity and utilization.
In this paper, we focus on online inference efficiency, {\em i.e.}, latency and memory usage, by introducing a task-aware expert merging framework. Unlike previous approaches that rely on fixed routing or need task tags and labeled datasets, our framework uses an online approach to deal with changing task distribution without requiring explicit task tags.

\textbf{\textit{Model Merging in Language Models}.}
Recent progress in model merging for language models shows that merging separately trained models can produce a single model with strong multi-task performance, without the need to access the original data or large computational resources~\cite{yang2024model}. Works~\cite{ilharco2023editing, yang2024adamerging} 
enable model merging by using task vectors that represent the parameter differences between pre-trained and fine-tuned models.
RegMean \cite{jin2023dataless} shows that for model merging, linear layers admit closed-form solutions based on training data statistics. Works \cite{matena2022merging, dk2024fisher} leverage the Fisher information matrix to estimate parameter importance during merging.
Instead of relying on fixed datasets or static merging strategies, we design a task-aware expert merging framework that can adaptively adjust merging decisions based on observed task distribution for online inference.

\textbf{\textit{Neural Bandit}.}
Neural contextual bandits use neural networks to model complex reward functions, allowing for both linear and non-linear relationships \cite{xu2022neural, qi2024robust}.
Authors in~\cite{pmlrzhou20a, zhang2021neural} provide regret guarantees for UCB- and Thompson Sampling-based algorithms using fully connected networks. EE-Net \cite{ban2022eenet} introduces an additional network to enhance exploration in neural bandits. 
Works \cite{wang2025neural} extend neural bandits to combinatorial actions, enabling the selection of multiple items.
In this paper, we frame the router design problem as optimizing a complex reward over a \textit{continuous decision space}. Unlike most prior neural bandit methods that focus on discrete actions, our framework handles a continuous, multi-dimensional decision space, allowing for flexible and adaptive expert merging during online inference. 
\section{System Model}\label{sec:model}



\vspace{-3mm}
\begin{figure}[htb]
	\centering
	\includegraphics[width=0.93\linewidth]{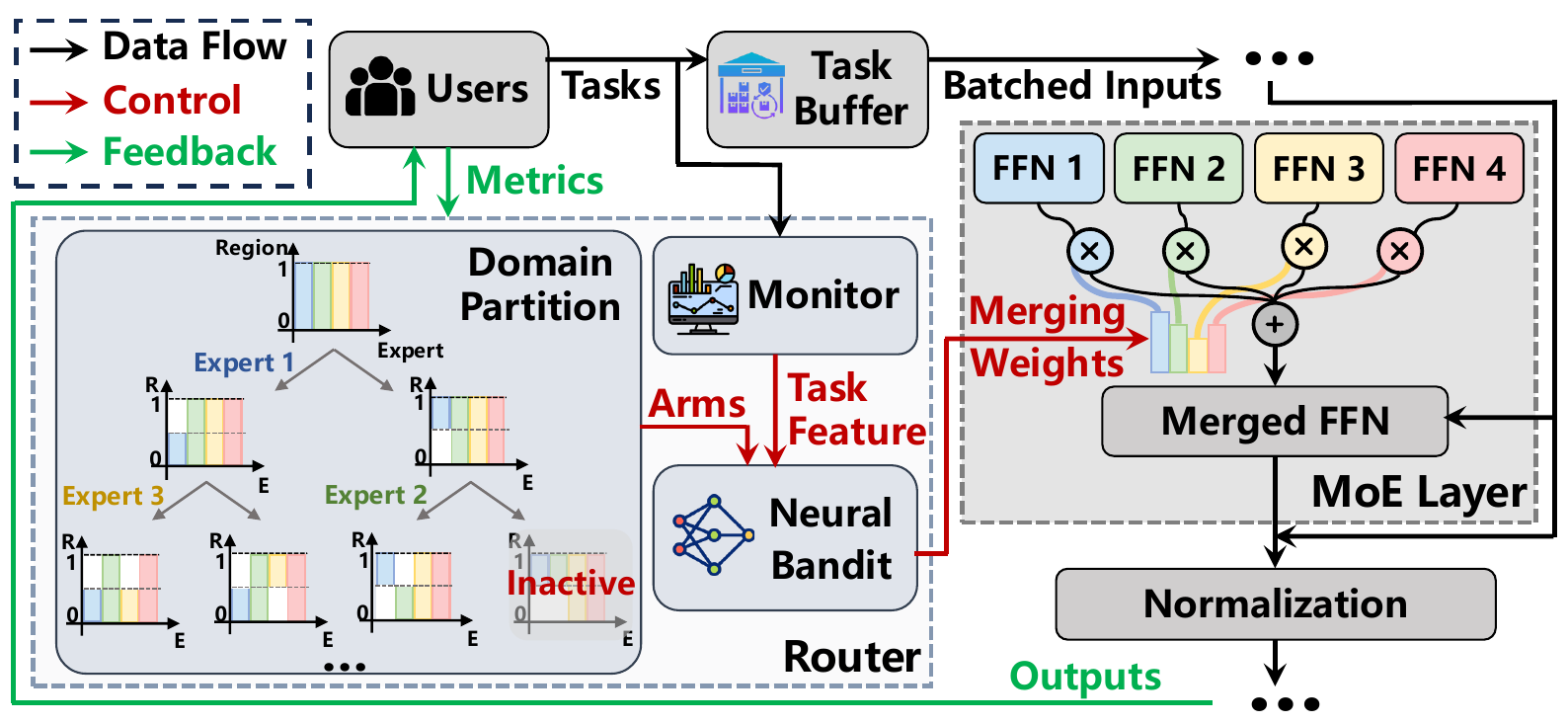}
	\vspace{-2mm}
	\caption{Overview of \texttt{Tanbr} architecture.}
	\label{fig:model}
\end{figure}
\vspace{-4mm}

\subsection{Preliminaries}
\textbf{\textit{Sparse Mixture of Experts}.} SMoE architectures enhance the computational efficiency and scalability of neural models, such as Transformers, by replacing feed-forward network (FFN) layers with MoE layers \cite{jacobs1991adaptive}. Each MoE layer is composed of {\small $K $} experts ({\small $K\geq2$}), with a router responsible for selecting the subset of experts for activation. It is assumed that all experts within a same layer share an identical FFN architecture \footnote{In cases of structural discrepancies, weight permutation alignment techniques can be employed to mitigate mismatches among neurons~\cite{li2024merge}.}.
For simplicity of presentation, we focus on a single MoE layer first, where each expert is parameterized as {\small $E(\cdot; \boldsymbol{W}_k),\forall k \in \mathcal{K}$}, with {\small $E: \mathbb{R}^d \rightarrow \mathbb{R}^d$} representing a single expert module, and {\small $\mathcal{K}$} denoting the set of {\small $K$} experts. 
For each input {\small $u$}, the router weights, denoted as {\small $\boldsymbol{x} = [x_k]_{k \in \mathcal{K}}^\top$} are used to linearly combine the outputs of the selected experts (with {\small $x_k = 0$}, meaning expert {\small $k$} is not activated). Specifically, the output is computed as: {\small $ \boldsymbol{y} = \sum_{k \in \mathcal{K}} x_kE(u; \boldsymbol{W}_k)$}.

Note that SMoE can operate at multiple levels, such as token \cite{fedus2022switch}, sequence \cite{zhong2024lory}, and task \cite{kudugunta2021beyond}. Specifically, expert selection can be based on a token {\small $c$} ({\small $\boldsymbol{x} = \text{Router}(c)$}), or a sequence of tokens {\small $[c_i]_{i\in \mathcal{I}}^\top$} ({\small$\boldsymbol{x} = \text{Router}(\frac{1}{I}\sum_{i \in \mathcal{I}} c_i)$}), or a task tag ({\small $\boldsymbol{x} = \text{Router}(\text{task\_tag})$}), where {\small $\text{Router}: \mathbb{R}^{V} \rightarrow \mathbb{R}^K$} denotes the router network/function. 
To improve training and inference efficiency, many studies use task-level routing strategies to train SMoE models \cite{kudugunta2021beyond,he2023merging} or use task-level routing SMoE by merging independently trained LLMs \cite{sukhbaatar2024branchtrainmix}.
However, SMoEs often suffer from load imbalance, and their computational cost increases significantly with the number of activated experts, thereby limiting scalability in real-world deployment.

\textbf{\textit{Expert Merging}.} To overcome the difficulty that discrete routing decisions hinder gradient backpropagation during training, expert merging methods are proposed, which compute a weighted combination of all FFNs \cite{muqeeth2023soft}.
Specifically, given an input {\small $u$}, the router outputs merging weights {\small $\boldsymbol{x}$}, which are used to merge expert parameters rather than expert selection. The output is:
{\small 
	$\boldsymbol{y} = E(u;\boldsymbol{W}), \boldsymbol{W} = {\textstyle \sum_{k \in \mathcal{K}}} x_{k}\boldsymbol{W}_k$
}.
Applying the merged expert for inference enables the system to leverage the ``combined" knowledge of multiple experts while keeping the computational cost of a single expert. Moreover, this method avoids uneven workload distribution across experts.

\subsection{Task-Aware Online Inference with MoE}
\textbf{\textit{Online System Model}.} 
As illustrated in Fig. \ref{fig:model}, we assume that the service provider is equipped with a pre-trained MoE model and aims to support {\small $V $} different types of tasks, where {\small $V \in \mathbb{Z}^+$}~\footnote{In this work, we assume that task-level routing was used during training, as it offers better compatibility. However, our method is still applicable even if token-level or sequence-level routing was used during training.}.
We focus on how to perform task-level routing during online inference because it can improve efficiency by reducing router activations and expert switching \cite{kudugunta2021beyond}.
When deployed for inference, the system operates in an \textit{online environment}, where requests arrive randomly.
Although each task is associated with a type, this information is typically unavailable to the router during online inference due to the absence of explicit task tags or labels.
Consequently, traditional task-level routing methods \cite{kudugunta2021beyond, he2023merging} that rely on this information are not effective.
To address this, we introduce a \textit{monitor} module (depicted in Fig. \ref{fig:model}), which estimates the current task distribution based on historical data and lightweight prediction tools \cite{gast2018lightweight, han2024inss}, such as exponential moving average, Count-Min Sketch \cite{cormode2005improved}. Specifically, at each time slot {\small $t$} (e.g., every 30 minutes or longer), the monitor generates a task feature vector {\small $\boldsymbol{\psi}_t = [\psi_{t,v}]_{v \in \mathcal{V}}^\top$}, where {\small $\psi_{t,v}$} denotes the estimated proportion of task type {\small $v$} during that slot.

\textbf{\textit{Core Issue}.} To further reduce computational and memory overhead during inference, we also adopt expert merging strategy to serve inference tasks.
The main challenge is to design and train a new router that can effectively leverage task features at each time slot to guide expert merging within the pre-trained MoE, without requiring explicit task tags. Note that our router must dynamically adapt to evolving task distributions to ensure efficient and reliable online inference.

\textbf{\textit{Router Learning Objective}.}
There are two situations:
\textbf{i) Fine-tune before online inference}: Although expert merging is resource-efficient, directly applying merged models for online inference can result in performance drops, especially for post-layer normalization architectures (e.g., T5-based MoEs \cite{raffel2020exploring}). They are sensitive to parameter shifts and therefore require fine-tuning before online adaptation.
This fine-tuning phase is performed with the labeled training data, on which the system can calculate task-specific losses (e.g., Mean Squared Error or Cross-Entropy) to fine-tune the model and train the router.
For each time slot {\small $t$}, the loss vector is denoted as {\small $\boldsymbol{l}_t = [l_{t,v}]_{v \in \mathcal{V}}^\top$}, where {\small $l_{t,v}$} represents the average loss of task type {\small $v$} during that time slot. During the fine-tuning phase, the goal is to minimize the overall loss. Therefore, we define the reward of the router as the negative value of the loss, {\em i.e.},
{\small
	$\boldsymbol{r}_t = - \boldsymbol{l}_{t}$}.
\textbf{ii) Online learning during inference}: Architectures that use pre-layer normalization (such as BERT-based MoEs \cite{devlin2019bert}) show stronger robustness when merging experts, allowing the merged model to be used directly for inference.
In this setting, the reward can be directly defined using performance metrics ({\em e.g.}, accuracy) observed during inference:
{\small $\boldsymbol{r}_t = \boldsymbol{a}_{t}, \boldsymbol{a}_t = [a_{t,v}]_{v \in \mathcal{V}}^\top$}, where {\small $a_{t,v}$} denotes the average reward of task type {\small $v$} during time slot {\small $t$}.
This removes the need for labeled data and allows the router to be trained fully in an online manner.
\textit{More generally}, the reward can be abstracted as a function of the merging weights {\small $\boldsymbol{x}_t$} at each time slot {\small $t$}, {\em i.e.},
{\small 
	$\boldsymbol{r}_t = f^*(\boldsymbol{x}_t)$,
}
where {\small $f^*: \mathbb{R}^{K} \rightarrow \mathbb{R}^{V}$} denotes the non-linear mapping from the merging weight to the reward, omitting intermediate steps such as expert merging, model execution, and loss/accuracy calculation.
The total reward is computed as the weighted sum {\small $\boldsymbol{\psi}_t^\top \boldsymbol{r}_t$}, which is linear in {\small $\boldsymbol{r}_t$}.

\subsection{Problem Formulation}
\textbf{\textit{Mathematical Formulation}.} 
We formulate the problem as an online constrained learning optimization problem, aiming to maximize the cumulative reward over a finite time horizon {\small $\mathcal{T}$}, as follows,

\vspace{-3mm}
{\small
	\begin{equation}
		\max {\textstyle \sum_{t \in \mathcal{T}} } \boldsymbol{\psi}_t^\top \boldsymbol{r}_t,
		\label{problem1}
\end{equation}}
\vspace{-7mm}
{\small
	\begin{align}
		\text{s.t.} \quad &{\textstyle \sum_{k \in \mathcal{K}}} x_{t,k}=1, \forall t \in \mathcal{T}. \tag{\ref{problem1}a}\\
		&{\textstyle \sum_{k \in \mathcal{K}}} \Pi(x_{t,k}>0) \leq B, \forall t \in \mathcal{T}, \tag{\ref{problem1}b}\\
		& \boldsymbol{x}_{t} \in \mathcal{X}, \forall t \in \mathcal{T}, \tag{\ref{problem1}c}
\end{align}}
\vspace{-5mm}

\noindent
where {\small $\mathcal{X}$} is a measurable space of merging weights, {\em i.e.}, {\small $\mathcal{X} = [0,1]^K$}; {\small $\Pi(\cdot)$} is an indicator function,  which returns {\small $1$} if the input condition is satisfied and {\small $0$} otherwise. 
Constraint (\ref{problem1}a) ensures that the sum of merging weights produced by the router is equal to {\small $1$}. 
Constraint (\ref{problem1}b) limits the number of selected experts to at most {\small $B$}, a predefined constant that reflects available computational and memory resources.

\textbf{\textit{Challenges}.} i) Problem (\ref{problem1}) can be reduced to the well-known Knapsack Problem \cite{kellerer2004multidimensional, pang2023eris}, classifying it as an NP-hard problem. 
ii) The decision variables are both multi-dimensional and continuous, leading to high computational costs when searching for an optimal solution.
iii) The reward acquisition is context-dependent and follows a non-linear mapping, which introduces additional complexities.
iv) To preserve valuable information and utilize the problem’s linear components, the reward is multi-dimensional, which increases learning complexity and requires balancing multiple objectives.
\section{Algorithm Design}\label{sec:algo}


\subsection{Main Idea}

Given the challenges of Problem (\ref{problem1}), reinforcement learning (RL) is less suitable, due to the absence of long-horizon state transitions~\cite{fu2026heterogeneous}.
We therefore adopt an MAB framework, which offers theoretical performance guarantees and greater interpretability compared to end-to-end learning methods\cite{bouneffouf2024tutorial, dai2024cost}.
However, existing MAB methods are inadequate due to the presence of both non-linear parameters and a continuous decision space. 
Here, we introduce a novel framework, tree-structured adaptive neural bandit router, \texttt{Tanbr}, for fine-tuning and online inference within a pre-trained MoE. \texttt{Tanbr} includes the following steps:\\
i) \textbf{Tree Partitioning.} An infinite binary cover tree is used to discretize the \textit{continuous decision space}, where each decision corresponds to a feasible merging weight.
The tree structure allows for the systematic generation of candidate merging weights that satisfy the given constraints. \\
ii) \textbf{Neural Bandit Framework.} To handle the \textit{non-linear dependencies} of reward, a neural bandit approach is employed. It utilizes the power of deep neural networks to learn the mapping from merging weights to rewards, while the UCB method \cite{pmlrzhou20a} guides the selection of merging weights by balancing exploration and exploitation. 
The network’s output layer generates \textit{multi-dimensional reward} estimates, allowing the merged expert to effectively support multiple tasks.\\
iii) \textbf{Incremental Learning and Exploration.} The selected decision guides the merging of experts within the pre-trained MoE for fine-tuning/inference, the observed reward is then used to update the parameters of \texttt{Tanbr}. As exploration progresses, the tree is incrementally expanded to generate a finer set of candidate merging weights for improved decision-making.

\subsection{Tree-Structured Measurable Space Partitioning}
\label{tree}
To effectively explore the continuous and multi-dimensional decision space of merging weights, we use a tree-structured partitioning method \cite{lazaric2014online}, which narrows promising regions while keeping the search complexity manageable. The tree structure also enables the pruning of subregions with unpromising merging weights, reducing unnecessary exploration.

\textbf{\textit{Partition Tree Design}.}
We adopt a pre-defined partition tree {\small $\mathcal{H}:= \{\mathcal{H}_{h,i}\}$}  over the measurable space {\small $\mathcal{X}$} to facilitate the optimization process. Each node {\small $\mathcal{H}_{h,i}$} represents a subregion within the space {\small $\mathcal{X}$} where {\small $h$} denotes the depth and {\small $i$} the index at that depth.
As illustrated in the \textit{domain partition} module in  Fig.~\ref{fig:model}, the tree recursively partitions the decision space in a binary and non-overlapping manner, following the rule: {\small $\mathcal{H}_{0,1} = \mathcal{X}$}, {\small $\mathcal{H}_{h,i} = \mathcal{H}_{h+1,2i-1} \cup \mathcal{H}_{h+1,2i}$}.
At the start of the algorithm, the tree {\small $\mathcal{H}$} is initialized with only the root node {\small $\mathcal{H}_{0,1} = \mathcal{X}$}.
At each time slot {\small $t$}, a candidate merging weight is selected from a discrete set sampled from all current \textit{active leaf nodes}.
The number of times node {\small $(h,i)$} has been selected up to time {\small $t$} is denoted as {\small $P_{h,i}(t)$}.
Once this count exceeds a depth-dependent threshold {\small $\tau_h(t)$}, the corresponding node is expanded into two child nodes, with the partitioning dimension randomly selected.
The threshold is defined as,  

{\small
\vspace{-2mm}
\begin{equation}
    \label{tau}
    \tau_h(t) = \frac{C^2\log(t/\delta)}{\nu_1^2}\rho^{-2h},
\end{equation}
\vspace{-4mm}
}

\noindent
where  {\small $C$}  is an fixed constant,  {\small $\nu_1$}  and  {\small $\rho$}  are smoothness parameters (as specified in Assumption \ref{smoothness}), and  {\small $\delta$}  is the confidence parameter.
Higher thresholds encourage more exploration before partitioning, while lower thresholds lead to earlier node expansion but may result in insufficient sampling. Adjusting the threshold parameter helps balance fine-grained partitioning with adequate exploration at each node.

\textbf{\textit{Constraint Satisfaction}.}
Given the set of active leaf nodes, candidate merging weights are generated by identifying points within their respective cover region that satisfy the \textit{constraint~(\ref{problem1}a)}. 
A leaf node is considered \textit{active} if its cover region contains at least one feasible merging weight; otherwise, it is marked \textit{inactive} and excluded from future selection and expansion.
For example, if a leaf node has a cover region {\small $[[0.5, 1], [0.75, 1]]$}, and no point within this region satisfies the constraint (\ref{problem1}a), the node is marked as inactive.
The search for feasible merging weights within a region can be formulated as a linear programming (LP) problem. This problem can be efficiently solved using standard optimization methods, such as the simplex method or interior-point algorithms~\cite{dantzig2016linear}.
To satisfy the constraint (\ref{problem1}b), the merging weights are adjusted by retaining the {\small $B$} largest values, normalizing them, and setting the rest to zero.

\subsection{Design of \texttt{Tanbr}}

\textbf{\textit{Neural Network for Reward Prediction}.} To capture the non-linear uncertainty in the reward function, we employ a fully connected neural network to learn and predict the reward function {\small $f^*(\boldsymbol{x})$}, and reformulate it as,

{\small 
\vspace{-2mm}
\begin{equation}
f(\boldsymbol{x},\boldsymbol{\theta}) = \sqrt{w}\boldsymbol{\theta}_L\sigma(\boldsymbol{\theta}_{L-1}\sigma(\dots \sigma(\boldsymbol{\theta}_1\boldsymbol{x}))),
\end{equation}}
\vspace{-5mm}

\noindent
where {\small $w$} is the width of each hidden layer, {\small $L$} denotes the depth of the network, and {\small $\sigma(\cdot)$} is the activation function. The output layer has a dimension of {\small $V$} to predict multi-dimensional rewards for the $V$ different task types.
The parameter vector of the network is defined as {\small $\boldsymbol{\theta} = [\boldsymbol{\theta}_1^\top, \dots, \boldsymbol{\theta}_L^\top]^\top \in \mathbb{R}^p$}, where {\small $p = wK+wV+w^2(L-1)$}.
Denote the gradient of {\small $f(\boldsymbol{x},\boldsymbol{\theta})$} as {\small $\boldsymbol{g}(\boldsymbol{x},\boldsymbol{\theta}) = \nabla_{\boldsymbol{\theta}} f(\boldsymbol{x},\boldsymbol{\theta}) \in \mathbb{R}^p$}.

Our proposed tree-structured adaptive neural bandit router, \texttt{Tanbr} is presented in Alg. \ref{algo_1}, as follows:

\textbf{\textit{Initialization}.}
\texttt{Tanbr} begins by initializing the neural network, where each element of the parameter vector {\small $\boldsymbol{\theta} $} is randomly generated from an appropriate Gaussian distribution \cite{pmlrzhou20a}. Furthermore, the partition tree is initialized with a root node that spans the entire decision space, {\em i.e.}, {\small $\mathcal{X} = [0, 1]^K$}.

\textbf{\textit{Decision Making}.} 
At each time slot {\small $t$}, \texttt{Tanbr} gets the task feature vector {\small $\boldsymbol{\psi}_t$} and extracts the set of active leaf nodes {\small $\mathcal{N}_t$} from the partition tree {\small $\mathcal{H}$}  (lines \ref{line:3}-\ref{line:4}).
For each active node, a feasible merging weight is generated that satisfies both its cover region and constraint (\ref{problem1}a), forming the candidate set (line \ref{line:5}). Then, for each merging weight {\small $\boldsymbol{x}_{h,i}$}, line~\ref{line:6} calculates its UCB {\small $U_{h,i}$} by,

\vspace{-4mm}
{\small 
\begin{align}
    \label{u_node}
    U_{h,i} = & \boldsymbol{\psi}_t^\top f(\boldsymbol{x}_{h,i}, \boldsymbol{\theta}_{t-1}) +\nu_1\rho^h + \nonumber\\ 
    & \gamma_{t-1} \sqrt{\boldsymbol{g}(\boldsymbol{x}_{h,i}, \boldsymbol{\theta}_{t-1})^\top \boldsymbol{Z}_{t-1}^{-1} \boldsymbol{g}(\boldsymbol{x}_{h,i}, \boldsymbol{\theta}_{t-1})/w},
\end{align}
}
\vspace{-5mm}

\noindent
where {\small $\boldsymbol{Z}$} is an auxiliary variable; 
{\small $\gamma_t$} is a positive scaling factor and  is defined as 
{\small $\gamma_t = \sqrt{1 + C_1 w^{-1/6} \sqrt{\log w} L^4 t^{7/6} \lambda^{-7/6}} \cdot (\upsilon \sqrt{\log (\det \boldsymbol{Z}_t/\det \lambda \boldsymbol{I}) + C_2 w^{-1/6} \sqrt{\log w} L^4 t^{5/3} \lambda^{-1/6} 
- 2 \log \delta} $
$+ \sqrt{\lambda} S)+ (\lambda + C_3 t L) [ (1 - \eta w \lambda)^{J/2} \sqrt{t/\lambda} + w^{-1/6}\sqrt{\log w} \cdot$
$ L^{7/2} t^{5/3} \lambda^{-5/3} (1 + \sqrt{t/\lambda})]$},
where  {\small $\lambda$}  denotes the regularization parameter, {\small $S$} is the norm parameter, and {\small $\eta$}  represents the step size for updating the parameters of the neural network.
The first term estimates the expected reward, the second reflects the granularity of the node’s region in the tree, and the third term quantifies the uncertainty in the reward prediction. 
Line~\ref{line:7} selects merging weight  $\boldsymbol{x}_{h_t,i_t}$  with the highest upper bound.

\textbf{\textit{Parameter Updating}.} The selected merging weight {\small $\boldsymbol{x}_{h_t,i_t}$} is used to merge experts, and the merged model is deployed for fine-tuning/inference.
Afterward, the corresponding reward {\small $f^*(\boldsymbol{x}_{h_t,i_t})$} is observed (line~\ref{line:8}).
\texttt{Tanbr} then updates the selection count {\small $P_{h,i}$} for all active leaf nodes in {\small $\mathcal{N}_t$} (line~\ref{line:9}).  If the selected node has been explored sufficiently, {\em i.e.}, the number of times the node has been selected {\small $P_{h_t,i_t}(t)$} exceeds the threshold {\small $\tau_{h_t}(t)$}, it is expanded (lines~\ref{line:10}-\ref{line:12}). As described in Sec.~\ref{tree}, the expansion is performed by randomly selecting a dimension of the region for binary partitioning, which results in two new leaf nodes being added to {\small $\mathcal{H}$}.
At the end of {\small $t$}, line~\ref{line:13} updates the auxiliary variable {\small $\boldsymbol{Z}_t$} and line~\ref{line:14} updates the neural network parameters {\small $\boldsymbol{\theta}_t$} by minimizing the loss function {\small $L(\boldsymbol{\theta})$} using stochastic gradient descent (SGD) with a step size {\small $\eta$}. The loss function we used is:
{\small
$L(\boldsymbol{\theta}) = {\textstyle \sum_{s=1}^{t}} \frac{1}{2} \left( f(\boldsymbol{x}_{h_s,i_s}, \boldsymbol{\theta}) - f^*(\boldsymbol{x}_{h_s,i_s}) \right)^2 + \frac{\lambda}{2} \|\boldsymbol{\theta} - \boldsymbol{\theta}_0\|_2^2$
}, where {\small $\lambda$} denotes the regularization parameter.

{\small
	\vspace{-2mm}
	\begin{algorithm}[htb]
		\caption{Tree-structured adaptive neural bandit router}
		\label{algo_1}
		\begin{algorithmic}[1]
			\State \textbf{Initialization:} $\boldsymbol{\theta}_0$, $\boldsymbol{Z}_0 = \lambda \boldsymbol{I}$, $\mathcal{H} =\{[0,1]^K\}$. \label{line:1}
			\For{$t \in \mathcal{T}$} \label{line:2}
			\State Get the task feature vector $\boldsymbol{\psi}_t$ from monitor; \label{line:3}
			\State Get the set of active leaf nodes $\mathcal{N}_t$ from the tree $\mathcal{H}$; \label{line:4}
			\State Generate the candidate set $\{\boldsymbol{x}_{h,i}\}_{(h,i) \in \mathcal{N}_t}$ using
			\Statex optimization methods; \label{line:5}
			\State Compute $U_{h,i}$ for all $\{\boldsymbol{x}_{h,i}\}_{(h,i) \in \mathcal{N}_t}$ using Eq. (\ref{u_node});\label{line:6}
			\State Let $(h_t,i_t) = \arg\max_{(h,i) \in \mathcal{N}_t} U_{h,i}$; \label{line:7}
			\State Apply $\boldsymbol{x}_{h_t,i_t}$ to merge experts for fine-tuning/
			\Statex inference  and then observe reward $f^*(\boldsymbol{x}_{h_t,i_t})$; \label{line:8}
			\State Update selection count: $P_{h_t,i_t}(t) = P_{h_t,i_t}(t-1)+1$; 
			\Statex $P_{h,i}(t) = P_{h,i}(t-1), \quad \forall (h,i) \in \mathcal{N}_t/(h_t,i_t)$; \label{line:9}
			\If {$P_{h_t,i_t}(t) \geq \tau_{h_t}(t)$} \label{line:10}
			\State $\mathcal{H} = \mathcal{H} \cup expand(h_t,i_t)$; \label{line:11}
			\EndIf \label{line:12}
			\State Update auxiliary variable with gradient:
			$\boldsymbol{Z}_t = \boldsymbol{Z}_{t-1} $
			\Statex $+  \quad \boldsymbol{g}(\boldsymbol{x}_{h_t,i_t}, \boldsymbol{\theta}_{t-1}) \cdot \boldsymbol{g}(\boldsymbol{x}_{h_t,i_t}, \boldsymbol{\theta}_{t-1})^\top/w$; \label{line:13}
			\State Update: $\boldsymbol{\theta}_t =\boldsymbol{\theta}_{t-1} - \eta \nabla L(\boldsymbol{\theta}_{t-1})$ with {\small $J$} steps; \label{line:14}
			\EndFor
		\end{algorithmic}
	\end{algorithm}
	\vspace{-2mm}}

\subsection{Theoretical Analysis}\label{sec:algo_analysis}

To facilitate a more concise and structured theoretical analysis, we augment the neural network with an additional linear output layer parameterized by {\small $\boldsymbol{\psi}_t$}. Let the complete set of parameters be denoted by {\small $\boldsymbol{\xi} = [\boldsymbol{\theta}^\top, \boldsymbol{\psi}_t^\top]^\top$}. Accordingly, the estimated and real total reward is expressed as {\small $f(\boldsymbol{x},\boldsymbol{\xi})$} and {\small $f_1^*(\boldsymbol{x})$}. All detailed proofs are provided in the Appendix.

\textbf{\textit{Candidate Arms Upper Bound}.}
We first write some assumptions \cite{lazaric2014online}.

\begin{Assumption} i) \textnormal{\textbf{(Dissimilarity)}} The space {\small $\mathcal{X}$} is equipped with a dissimilarity function {\small $\ell : \mathcal{X} \times \mathcal{X} \to \mathbb{R}$} such that {\small $\ell(\boldsymbol{x}, \boldsymbol{x}') \geq 0$} for all {\small $\boldsymbol{x}, \boldsymbol{x}' \in \mathcal{X}$}, and {\small $\ell(\boldsymbol{x}, \boldsymbol{x}) = 0$}.
ii) Define the diameter of a subset {\small $\mathcal{A} \subseteq \mathcal{X}$} as 
{\small $diam(\mathcal{A}) = \sup_{\boldsymbol{x}, \boldsymbol{y} \in \mathcal{A}} \ell(\boldsymbol{x}, \boldsymbol{y})$}, and the {\small $\ell$}-open ball of radius {\small $\epsilon > 0$} and center {\small $\boldsymbol{x} \in \mathcal{X}$} as 
{\small $\mathcal{B}(\boldsymbol{x}, \epsilon) = \{\boldsymbol{x}' \in \mathcal{X}: \ell(\boldsymbol{x}, \boldsymbol{x}') \leq \epsilon\}$}.
iii) \textnormal{\textbf{(Local Smoothness)}}
\label{smoothness}
We assume that there exist constants {\small $\nu_1, \nu_2 > 0$} and {\small $0 < \rho < 1$} such that for all nodes {\small $(h, i)$}:
a) {\small $diam(\mathcal{H}_{h,i}) \leq \nu_1 \rho^h$};
b) There exists {\small $\boldsymbol{x}_{h,i}^o \in \mathcal{H}_{h,i}$} such that: 
{\small $\mathcal{B}_{h,i} = \mathcal{B}(\boldsymbol{x}_{h,i}^o, \nu_2 \rho^h) \subseteq \mathcal{H}_{h,i}$};
iii) {\small $\mathcal{B}_{h,i} \cap \mathcal{B}_{h,j} = \emptyset$} for all {\small $i \neq j$},
iv) For all {\small $\boldsymbol{x} \in \mathcal{X}$}, {\small $f^* - f(x) \leq \ell(\boldsymbol{x}^*, \boldsymbol{x})$}.
\end{Assumption}
\vspace{-2mm}

The dissimilarity assumption measures the difference between any two points in space. The local smoothness assumption ensures that the space is smoothly partitioned: regions shrink as depth increases, are non-overlapping, and the function behaves smoothly within each region.

\begin{Lemma}
	\label{lemma:N}
    Given the threshold {\small $\tau_h(t)$} defined in Eq. (\ref{tau}), the number of candidate merging weights at {\small $\forall  t$} is bounded by: 
    {\small
    \vspace{-1mm}
    \begin{equation}
        N_t \leq N = 2^{1/(1-\rho)} T \nu_1^2/\left(C^2 \log(T/\delta)\right) = \mathcal{O}(T).
\end{equation}
}
\end{Lemma}

\textbf{\textit{Regret Upper Bound}.}
The regret of an algorithm is defined as the difference between the total expected reward achieved by the optimal offline solution and that obtained by
the algorithm. Let {\small $x_t^*$} be the merging weight with the highest reward, {\em e.g.}, {\small $x_t^* = \boldsymbol{\psi}_t\arg\max_{\boldsymbol{x} \in \mathcal{X}}f^*(\boldsymbol{x})$}.
The regret of \texttt{Tanbr} is defined as follows:
{\small
    $R = {\textstyle \sum_{t \in \mathcal{T}}} R_t, \quad R_t = \mathbb{E}\left[ f_1^*(\boldsymbol{x}_t^*) - f_1^*(\boldsymbol{x}_{h_t,i_t})\right]$}. 
Let {\small $\mathcal{X}_T = [\boldsymbol{x}_{h,i}]_{(h,i) \in \mathcal{N}_t, t \in \mathcal{T}}^\top$} be the complete set of candidate merging weights selected up to time {\small $T$}. By Lemma~\ref{lemma:N}, we have {\small $|\mathcal{X}_T| \leq TN$}.
We first introduce the following assumption.

\begin{Assumption}
\label{ass_1}
For any {\small $\boldsymbol{x} \in \mathcal{X}_T$}, it satisfies {\small $\|\boldsymbol{x}\|_2 = 1$} and {\small $[\boldsymbol{x}]_j = [\boldsymbol{x}]_{j + V/2}, \text{for } 1 \leq j \leq V/2$}.
Furthermore, for some {\small $\lambda_0 > 0$}, the NTK matrix {\small $\boldsymbol{M}$} satisfies {\small $\boldsymbol{M} \succeq \lambda_0 \boldsymbol{I}$}.
\end{Assumption}

This assumption is commonly used in neural bandit research~\cite{pmlrzhou20a}. The constraint {\small $\|\boldsymbol{x}\|_2 = 1$} is introduced to normalize the context vector. The symmetry condition enables the construction of an equivalent context vector in the form {\small $\boldsymbol{x}' = [\boldsymbol{x}^\top, \boldsymbol{x}^\top]^\top/\sqrt{2}$}.

\begin{Theorem}
\label{thm:regret}
    Let {\small $\tilde{d}$} be the effective dimension of the NTK matrix {\small $\boldsymbol{M}$} 
    and {\small $\boldsymbol{m} = [f_1^*(\boldsymbol{x})]_{\boldsymbol{x} \in \mathcal{X}_T}^\top$} as the true reward vector of the candidate merging weights.
    There exist constants {\small $C_1, C_2 > 0$} such that for any {\small $\delta \in (0, 1)$}, if {\small $w \geq \operatorname{poly}(T, L, K, \lambda^{-1}, \lambda_0^{-1}, S^{-1}, \log(1/\delta))$},
{\small $\eta \leq C_1 (TdL + d\lambda)^{-1}$}, {\small $\lambda \geq \max\{1, S^{-2}\}$}, and {\small $S \geq \sqrt{2\boldsymbol{m}^\top \boldsymbol{M}^{-1} \boldsymbol{m}}$}, then with probability at least {\small $1 - \delta$}, the regret of Alg.~\ref{algo_1} satisfies,

\vspace{-3mm}
{\small \begin{align}
    R \leq & 3\sqrt{T}\sqrt{\tilde{d} \log(1 + T N / \lambda)+2} \cdot \nonumber \\
    & \left[\upsilon\sqrt{\tilde{d} \log(1 + T N / \lambda)+2-2\log \delta} \right. \nonumber \\
    & \left. +(\lambda+C_3TL)(1-\eta w\lambda)^{J/2}\sqrt{T/\lambda}+2\sqrt{\lambda S} \right]+T+1.
\end{align}}
\end{Theorem}
\vspace{-2mm}

\noindent
\textbf{Remark.} Following Remark 4.7 in~\cite{pmlrzhou20a}, by setting {\small $J = \mathcal{O}(T L / \lambda)$} we have {\small $(\lambda+C_3TL)(1-\eta w\lambda)^{J/2}\sqrt{T/\lambda} \leq \sqrt{\lambda S}$}.
By treating {\small $\nu$}, {\small $\lambda$}, and {\small $\eta$} as constants and applying Lemma~\ref{lemma:N}, the regret bound in Theorem~\ref{thm:regret} simplifies to {\small $R_T = \mathcal{O}(\tilde{d} \sqrt{T} \log(T))$}, matching the regret bounds in prior neural bandit studies \cite{pmlrzhou20a, zhang2021neural,xu2022neural, qi2024robust}. 
Compared to LinUCB \cite{chu2011contextual}, this bound includes an additional factor $\tilde{d}$, reflecting the complexity of neural representations. This extra cost is acceptable given the ability to capture more expressive and nonlinear context.
A key contribution of \texttt{Tanbr} is maintaining these theoretical guarantees in a continuous decision space. This is achieved through smoothness assumptions and adaptive partitioning techniques that ensure efficient exploration while avoiding exhaustive search.
This result formally shows the theoretical suitability of \texttt{Tanbr} for long-term deployment in online MoE systems.

\section{Performance Evaluation}\label{sec:experiment}

\subsection{System Implementation}
\textbf{\textit{MoE Models and datasets}.} 
We evaluate our \texttt{Tanbr} using both i) encoder-decoder, and ii) encoder-only Transformer models to ensure a comprehensive evaluation. 
For both models, every FFN layer is replaced by an MoE layer containing {\small $K=8$} experts with the same architecture.
The encoder-decoder model is based on T5 \cite{raffel2020exploring}, which includes {\small $12$} FFN layers in both the encoder and decoder, and has about {\small $1.0B$} parameters. The decoder-only model is based on BERT \cite{devlin2019bert} and contains {\small $12$} FFN layers with approximately {\small $660M$} parameters. Both models are pre-trained using task routing \cite{kudugunta2021beyond,sukhbaatar2024branchtrainmix}.

\textbf{\textit{Datasets}.} We evaluate the models on {\small $V = 8$} common tasks from the GLUE benchmark, including natural language inference ({\em i.e.}, MNLI, QNLI, RTE), sentiment analysis ({\em i.e.}, SST-2), sentence similarity ({\em i.e.}, MRPC, QQP, STS-B), and linguistic acceptability ({\em i.e.}, CoLA) \cite{wang2018glue, han2025hilora}.
Among them, CoLA is evaluated by the Matthews correlation coefficient, STS-B is evaluated by the average value of Pearson and Spearman correlation coefficients, and the remaining tasks are evaluated by accuracy. Each dataset is divided into training and validation sets in an {\small $80/20$} split.
Both the batch size and the maximum sequence length (for source and target) are set to {\small $128$}.

\textbf{\textit{Framework Setting}.} In \texttt{Tanbr}, we set the smoothness parameter {\small $\nu_1 = 1$}, {\small $\rho$} over {\small $\{0.3,0.5,0.7,0.9\}$}. The regularization parameter is {\small $\lambda = 1$}, and the exploration parameter is {\small $\upsilon = 1$}. The confidence level is set as {\small $\delta = 1$}. We use learning rate {\small $\eta$} over {\small $\{1e-4,1e-3,1e-2, 1e-1\}$}. The neural network has width {\small $d = 64$} and depth {\small $L = 2$}. The maximum number of experts selected for merging is fixed to {\small $B = 8$}. All experiments are run with PyTorch on NVIDIA A100 GPUs with {\small $40GB$} of memory.

\textbf{\textit{Baselines}.} 
We compare \texttt{Tanbr} with the following methods: 
i) \texttt{RL} \cite{clark2022unified}: a RL approach for expert selection in sparse MoE, 
assigning one expert per token. 
ii) \texttt{Switch} \cite{fedus2022switch}: a neural network-based method that learns to select one expert per token for sparse MoE routing. 
iii) \texttt{SMEAR} \cite{muqeeth2023soft}: a neural network-based token-level expert merging method.
iv) \texttt{NUCB} \cite{pmlrzhou20a}:  a neural bandit method that randomly samples a set of {\small 20} feasible merging weights and selects one by neural UCB at each step.
v) \texttt{Regmean} \cite{jin2023dataless}: a model merging method that computes weights using a closed-form solution to the merging problem. 
vi) \texttt{Fisher} \cite{dk2024fisher}: a model merging method that leverages Fisher information to estimate parameter importance and performs weighted merging accordingly.  
vii) \texttt{Average} \cite{singh2020model}: a model merging method that performs element-wise averaging of all parameters.

\subsection{Results with Fine-tuning and Inference}

\begin{figure}[htb]
	\centering
	\begin{tabular}{cc}
		\hspace{-4mm}
		\begin{minipage}[t]{0.5\linewidth}
			\centering            \includegraphics[width=.9\textwidth]{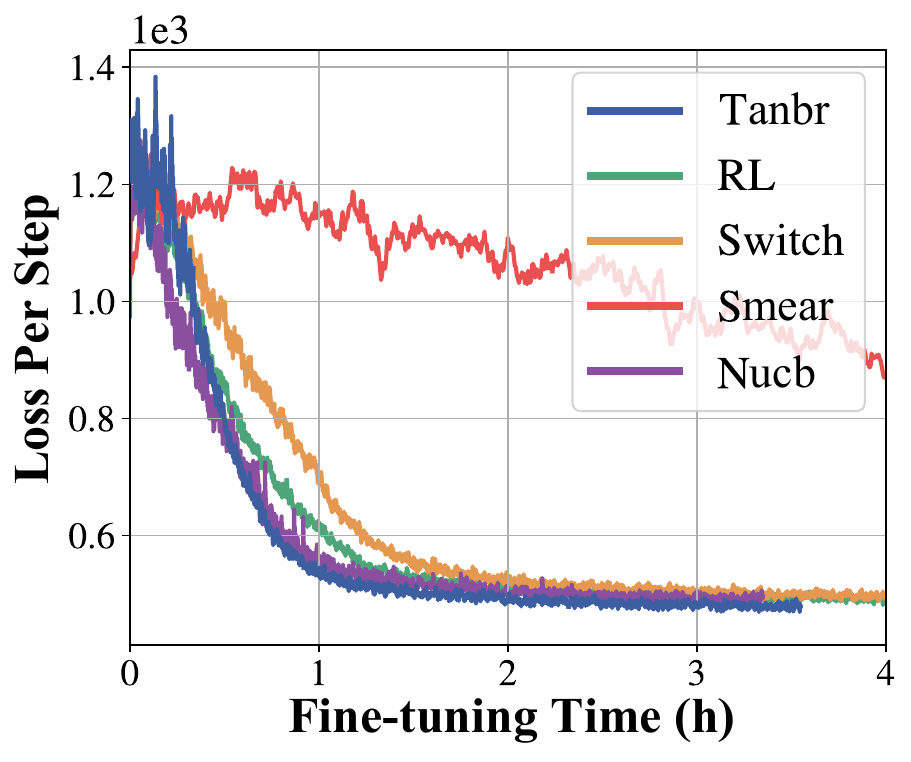}	
			\vspace{-1.5mm}
			\captionsetup{font=footnotesize}
			\caption*{(a) Fine-tuning Loss}
			\label{fig:loss_time}
		\end{minipage}
		\hspace{-2mm}
		\begin{minipage}[t]{0.5\linewidth}
			\centering
			\includegraphics[width=1.0\textwidth]{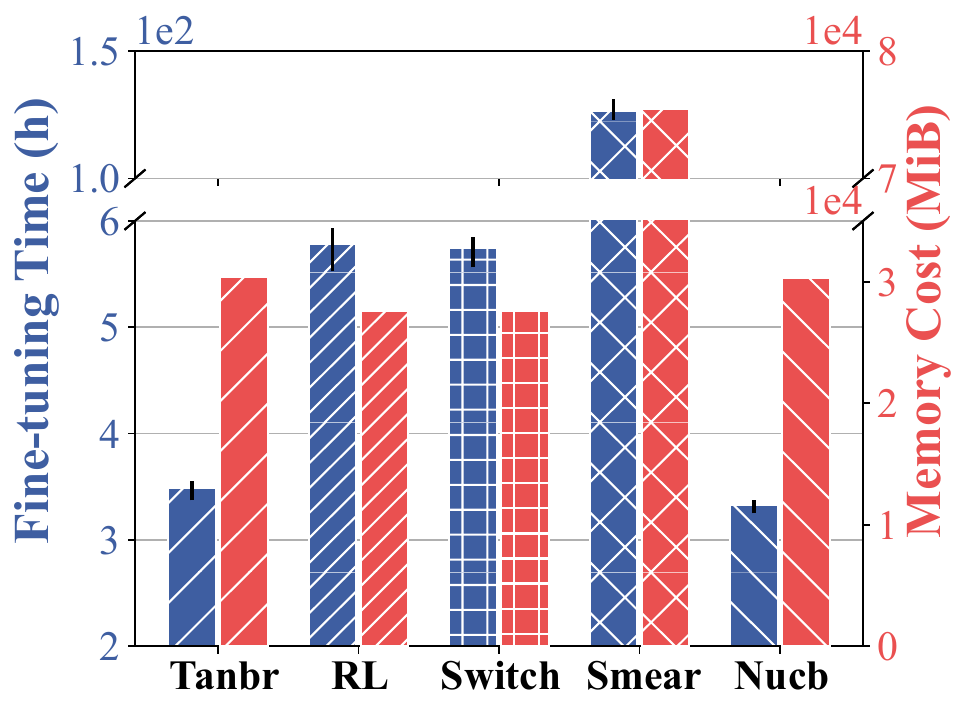}	
			\vspace{-6mm}
			\captionsetup{font=footnotesize}
			\caption*{(b) Time/Memory Cost}
			\label{fig:finetune_t5}
		\end{minipage}
	\end{tabular}
	\vspace{-2mm}
	\captionsetup{font=normalsize}
	\caption{Fine-tuning performance with T5-based MoE.
	}
	\label{fig:t5_finetune}
	\vspace{-4mm}
\end{figure}

\textbf{\textit{Fine-tuning Phase}.} As mentioned in Section~\ref{sec:model}, T5-based MoE with post-layer normalization requires fine-tuning when expert merging is applied due to its sensitivity to parameter shifts. Therefore, we fine-tune its normalization layers for {\small $10,000$} steps using the training data, while simultaneously training a new router with different algorithms.
The results are depicted in Fig.~\ref{fig:t5_finetune}, showing the training loss curve, total training time, and memory usage.
From the figure, we have some observations.
\textit{First}, \texttt{Tanbr} starts with a higher loss since it begins with fewer candidate merging weights. However, as the partition tree refines over time, \texttt{Tanbr} converges much faster and achieves lower final loss than all baselines. In comparison, \texttt{Nucb}, which selects decisions from randomly generated merging weights, exhibits more fluctuation and slower convergence. 
 \textit{Second}, both \texttt{Tanbr} and \texttt{Nucb} (task-level routing) achieve the \textit{smallest training time}, achieving approximately {\small $40\%$} reduction compared to \texttt{RL} and \texttt{Switch}. Although \texttt{Tanbr} introduces a small amount of overhead due to maintaining the partition tree, it remains highly efficient. In contrast, \texttt{Smear}, which merges at the token level, results in a {\small $20\times$} increase in computational cost because it cannot use \textit{lightweight adapters} under the given pre-trained MoE architecture, making its overhead unavoidable.
 \textit{Third}, 
 \texttt{Tanbr} and \texttt{Nucb} consume about {\small $10\%$} more memory than \texttt{RL} and \texttt{Switch}, as they need to maintain the loss gradients of multiple experts. \texttt{Smear} performs token-level merging, incurring the highest memory footprint, exceeding other methods by {\small $179\%$} because it stores token-wise loss gradients for all experts.

\vspace{-4mm}

\vspace{-2mm}
\begin{figure}[htb]
    \begin{tabular}{cc}
		\hspace{-4mm}
		\begin{minipage}[t]{0.5\linewidth}
			\centering            \includegraphics[width=.9\textwidth]{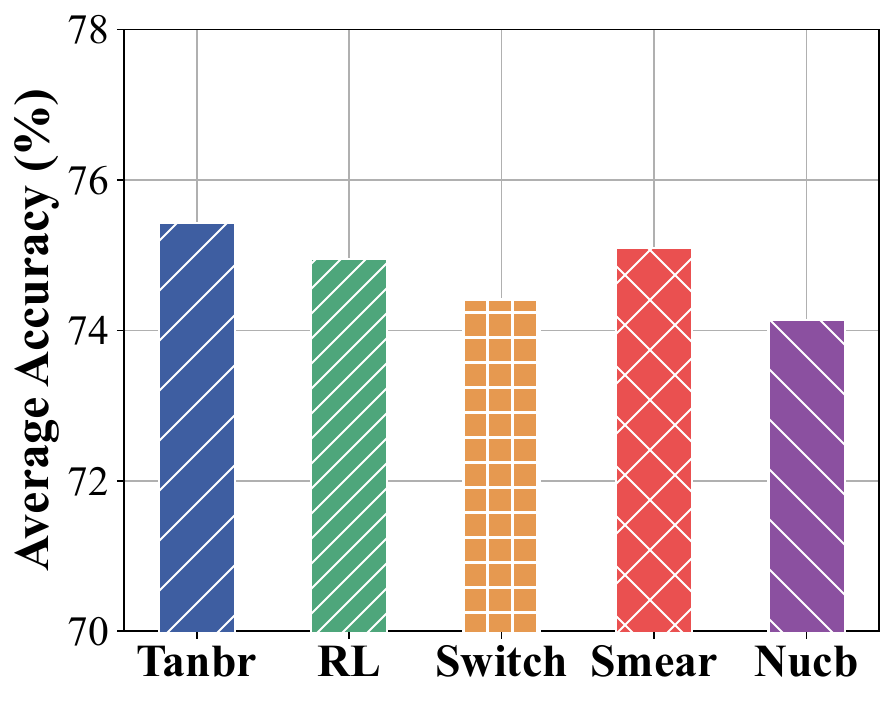}	
			\vspace{-1.5mm}
            \captionsetup{font=footnotesize}
            \caption*{(a) Average Accuracy}
			\label{fig:metric_t5}
		\end{minipage}
		\hspace{-2mm}
		\begin{minipage}[t]{0.5\linewidth}
			\centering
			\includegraphics[width=1.0\textwidth]{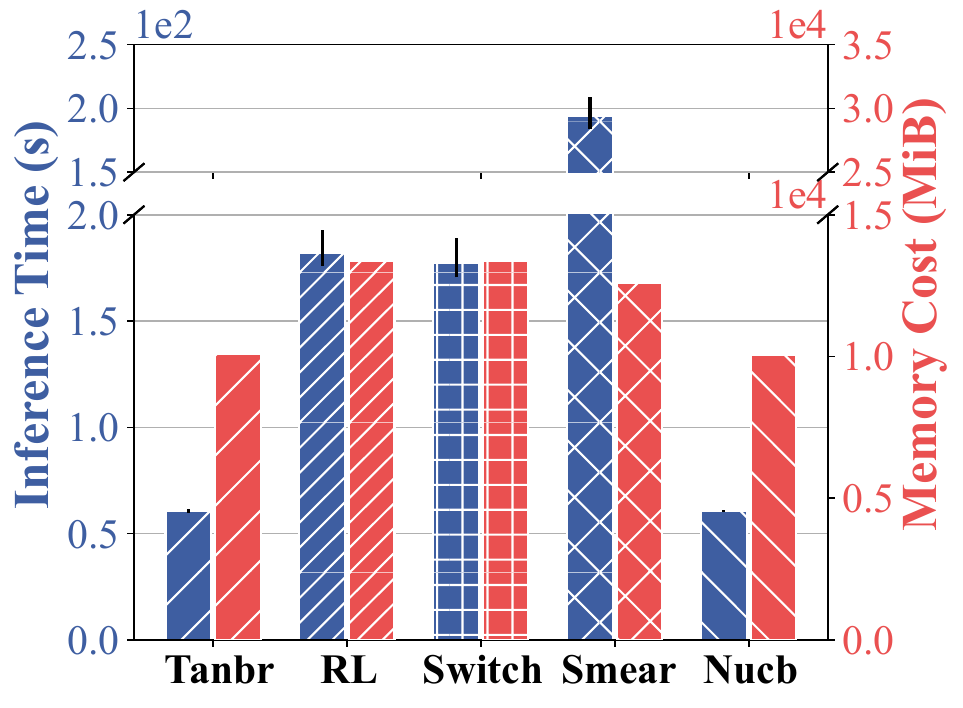}	
			\vspace{-5.5mm}
            \captionsetup{font=footnotesize}
            \caption*{(b) Time/ Memory Cost}
			\label{fig:inference_t5}
		\end{minipage}
	\end{tabular}
    \vspace{-2mm}
    \captionsetup{font=normalsize}
    \caption{Inference performance with T5-based MoE.
    }
    \label{fig:t5_inference}
    \vspace{-3mm}
\end{figure}

{\small
\begin{table*}[htbp]
	\centering
	\small
	\caption{Performance metrics for T5-based MoE with different routing methods.}
	\label{tab:taskwise_t5}
	\vspace{-2mm}
	{\footnotesize
	\begin{tabular}{l c c c c c c c c c}
		\toprule
		Method  & CoLA & MNLI & MRPC  & QNLI & QQP & RTE & SST-2 & STS-B   & Avg \\
		\ & mat & acc & acc/f1 & acc & acc/f1 & acc & acc & pear/spear\\
		\midrule
		\texttt{Tanbr}  & 42.9  & 75.3  & 78.4/84.4  & 84.1 & 82.0/74.8  & 63.8  & 91.5  & 86.1/85.9  & \textbf{75.43}  \\
		\midrule
		\texttt{RL}     & 42.1  & 75.7  & 77.0/83.3  & 82.6  & 81.6/74.2   & 65.3  & 91.2  & 83.6/83.8 & 74.95 \\
		\midrule
		\texttt{Switch}   & 43.2  & 71.5  & 79.3/85.6  & 82.8  & 81.6/75.2  & 62.3 & 90.4  & 84.3/84.1  & 74.41\\
		\midrule
		\texttt{Soft} & 43.8  & 74.0  & 79.9/85.9  & 82.9  & 81.5/74.6  & 62.8 & 90.6  & 85.8/85.7  & 75.10\\
		\midrule
		\texttt{Nucb}   & 40.4  & 73.2  & 80.8/84.6  & 82.6  & 81.9/74.3  & 62.8 & 90.3  & 83.3/82.8  & 74.14 \\
		\bottomrule
	\end{tabular}}
\vspace{-3mm}
\end{table*}}

\textbf{\textit{Online Inference}.} 
The fine-tuned models are deployed for online inference without further updates to the model parameters or router.
The inference results are presented in Fig.~\ref{fig:t5_inference}, including the average performance across tasks (detailed per-task metrics are provided in Tab. \ref{tab:taskwise_t5}), inference time per batch, and memory usage.
We highlight three main observations:
\textit{First}, \texttt{Tanbr} achieves the \textit{best average performance} along with consistently better performance on individual tasks.
This is because it dynamically adjusts the expert merging strategy based on online task features and fully exploits the information from multiple experts.
\textit{Second}, \texttt{Tanbr} demonstrates highly \textit{efficient inference speed}, requiring only one expert merging decision and execution at the start of each time slot. This reduces inference time per batch by {\small $65\%$} compared to \texttt{RL} and \texttt{Switch}, nearly {\small $99\%$} over \texttt{Smear}.
\textit{Third}, \texttt{Tanbr} achieves the \textit{lowest memory consumption}. Since only the merged model is loaded, 
it reduces memory use by {\small $20\%$} to {\small $25\%$}.

\subsection{Results with Online Learning for Inference}

\vspace{-4mm}
\begin{figure}[htb]
	\centering
	\begin{tabular}{cc}
		\hspace{-3mm}
		\begin{minipage}[t]{0.5\linewidth}
			\centering            \includegraphics[width=.95\textwidth]{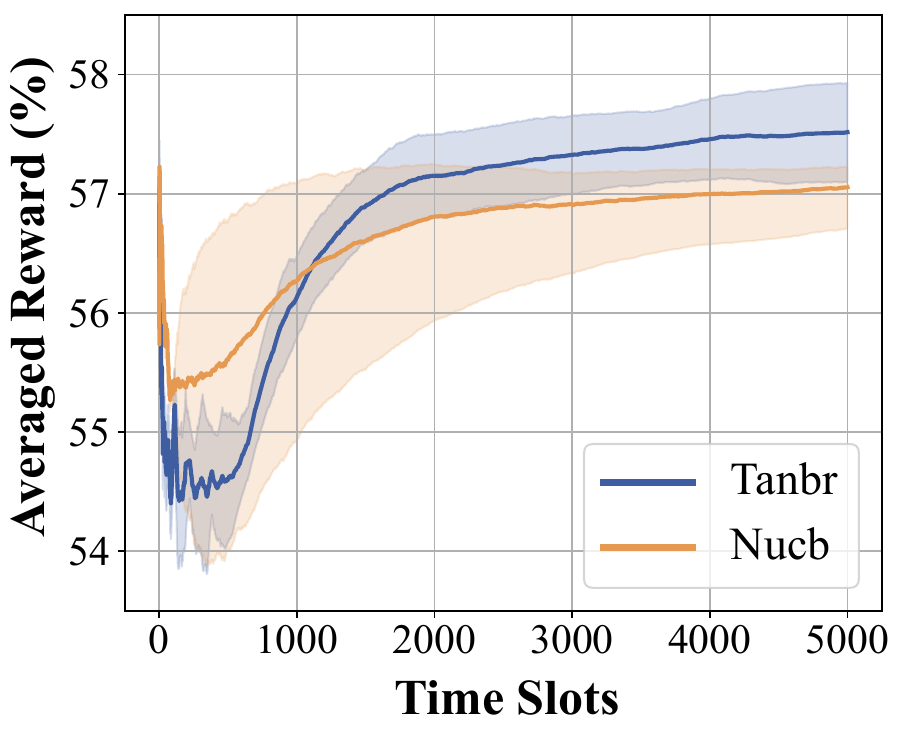}	
			\vspace{-1mm}
			\captionsetup{font=footnotesize}
			\caption*{(a) Online Reward}
			\label{fig:reward_time}
		\end{minipage}
		\hspace{-2mm}
		\begin{minipage}[t]{0.5\linewidth}
			\centering
			\includegraphics[width=1\textwidth]{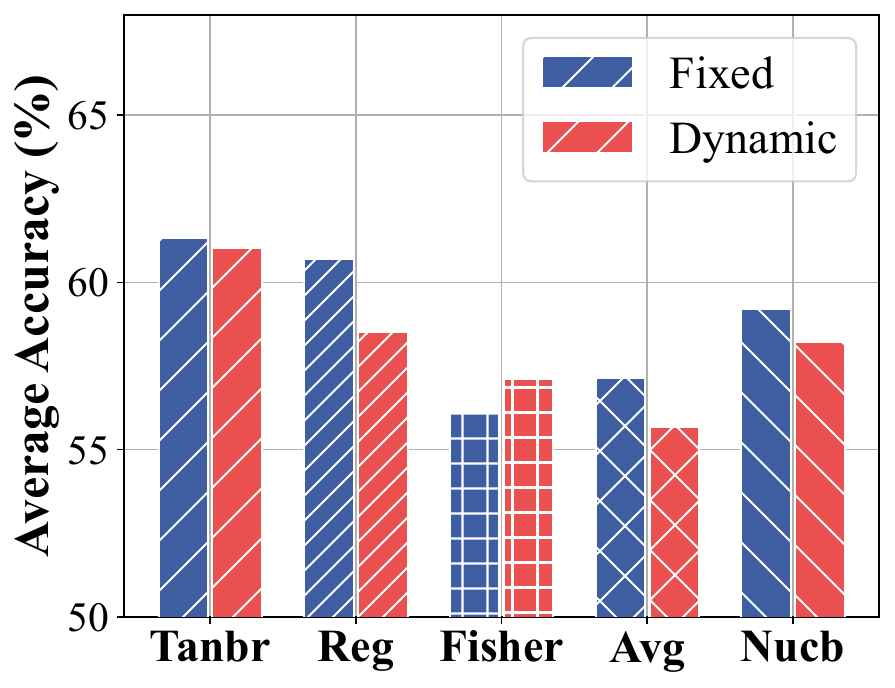}	
			\vspace{-6mm}
			\captionsetup{font=footnotesize}
			\caption*{(b) Average Accuracy}
			\label{fig:bert_metric}
		\end{minipage}
	\end{tabular}
	\vspace{-2mm}
	\captionsetup{font=normalsize}
	\caption{Online Inference with BERT-based MoE.
	}
	\label{fig:bert}
	\vspace{-3mm}
\end{figure}

Here, we focus on BERT-based MoE models that do not require fine-tuning and allow the router to be trained online during inference. Since loss information is unavailable during online inference, traditional router training methods cannot be applied. Thus, we compare our \texttt{Tanbr} with model merging methods.
The online learning performance is shown in Fig.~\ref{fig:bert}(a). Although \texttt{Tanbr} performs worse in the early stages due to the limited number of candidate merging weights, it converges faster and achieves better performance than \texttt{Nucb} as the partition tree becomes more refined over time. 
This result highlights the advantage of our tree partitioning strategy, which efficiently explores the search space during online updates. 
Overall, \texttt{Tanbr} demonstrates strong online learning capability.

\begin{figure}[!htb]
	\vspace{-4mm}
	\begin{tabular}{cc}
		\hspace{-3mm}
		\begin{minipage}[t]{0.48\linewidth}
			\centering
			\includegraphics[width=.95\textwidth]{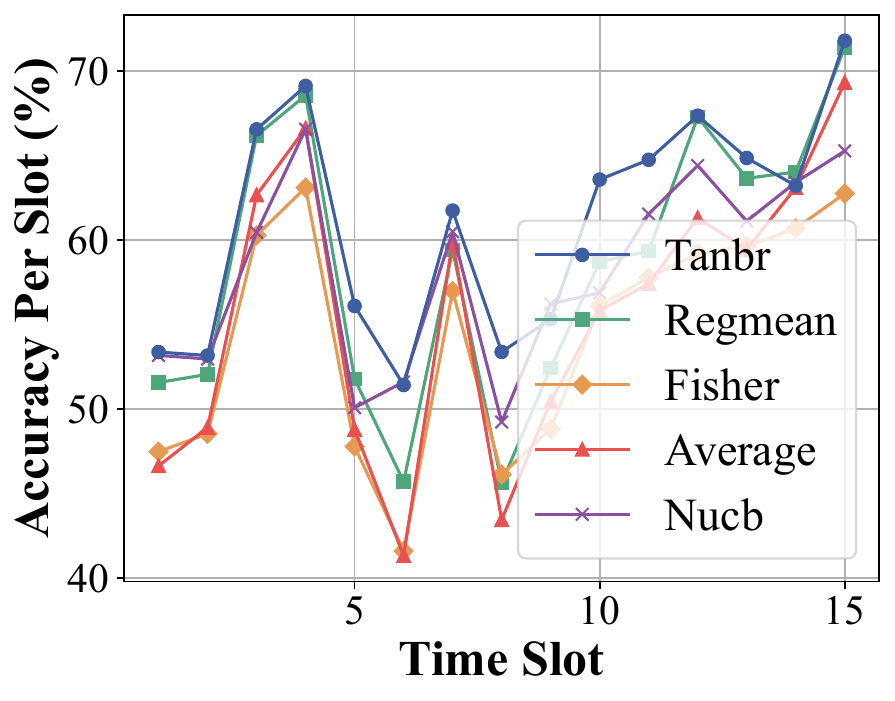}	
			\vspace{-2mm}
			\caption{Online inference per slot performance with BERT-based MoE.}
			\label{fig:ber_per_slot}
		\end{minipage}
		\hspace{0.3mm}
		\begin{minipage}[t]{0.48\linewidth}
			\centering
			\includegraphics[width=1\textwidth]{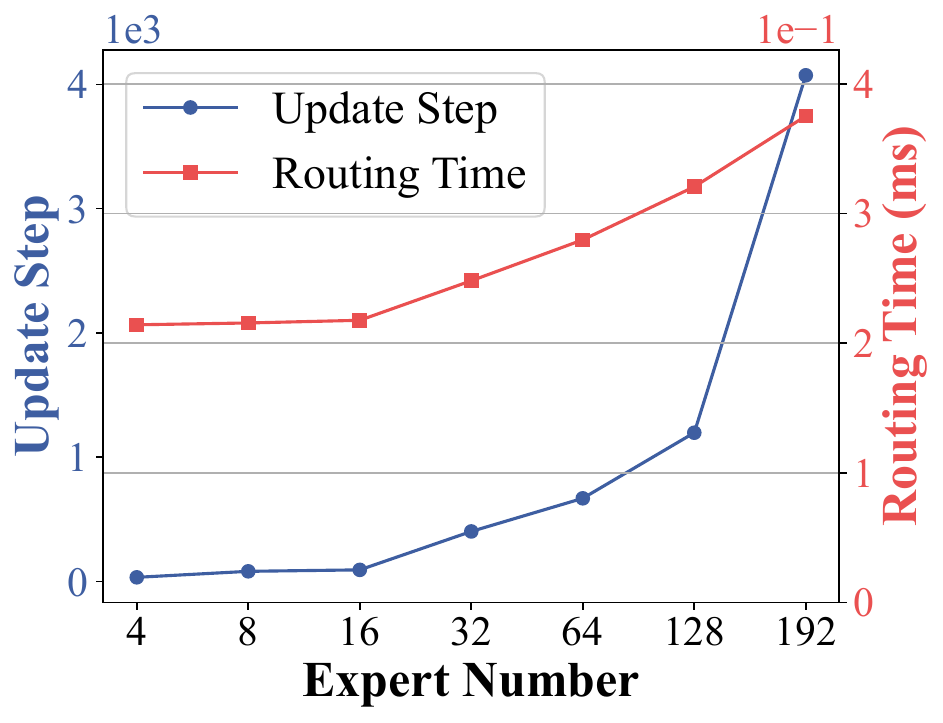}	
			\vspace{-6mm}
			\caption{Update and routing overhead of \texttt{Tanbr} under different system scales.}
			\label{fig:router_cost}
		\end{minipage}
	\end{tabular}
	\vspace{-3mm}
\end{figure}

After {\small $5000$} time slots of online learning, we further evaluate the performance over additional time slots. The results under both fixed and dynamic task features are shown in Fig.~\ref{fig:bert}(b) and the per-slot performance details under dynamic task features are shown in Fig.~\ref{fig:ber_per_slot}.
We can observe that \texttt{Tanbr}
outperforms the baselines by up to {\small $9\%$} under fixed task features and by {\small $5-10\%$} under dynamic task features.
The performance gain is especially notable in dynamic settings, as \texttt{Tanbr} adapts to evolving task distributions in real-time by merging experts based on task features. This adaptability enables it to surpass other methods that rely on static or less flexible merging strategies.

\subsection{Scalability and Ablation study}

\textbf{\textit{Scalability Evaluation}.} To evaluate the scalability of \texttt{Tanbr}, we conduct experiments on MoE models with 4 to 192 experts. Fig.~\ref{fig:router_cost} shows the number of update steps required to identify 20 feasible leaf nodes and the time needed to compute the optimal merging weight from them. The results show that as the number of experts increases, more update steps are required due to the higher dimensionality of the decision space {\small $\mathcal{X}$}, which leads to more partitioning options. Although the routing time per decision also increases with the number of experts, it remains low (less than 0.4 ms even when {\small $K = 192$}). These results indicate that \texttt{Tanbr} maintains high efficiency and scalability in large-scale settings.

\vspace{-4mm}
\begin{figure}[htb]
	\centering
	\begin{tabular}{cc}
		\hspace{-3mm}
		\begin{minipage}[t]{0.5\linewidth}
			\centering            \includegraphics[width=1\textwidth]{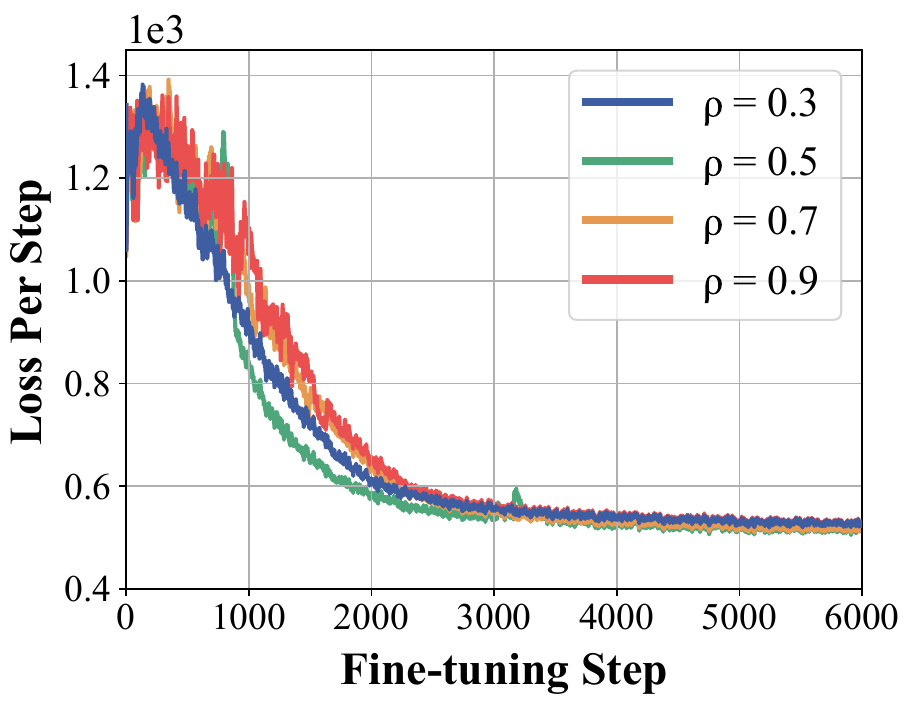}	
			\vspace{-6mm}
			\captionsetup{font=footnotesize}
			\caption*{(a) Smoothness Parameter {\small $\rho$}}
			\label{fig:rho}
		\end{minipage}
		\hspace{-2mm}
		\begin{minipage}[t]{0.5\linewidth}
			\centering
			\includegraphics[width=1\textwidth]{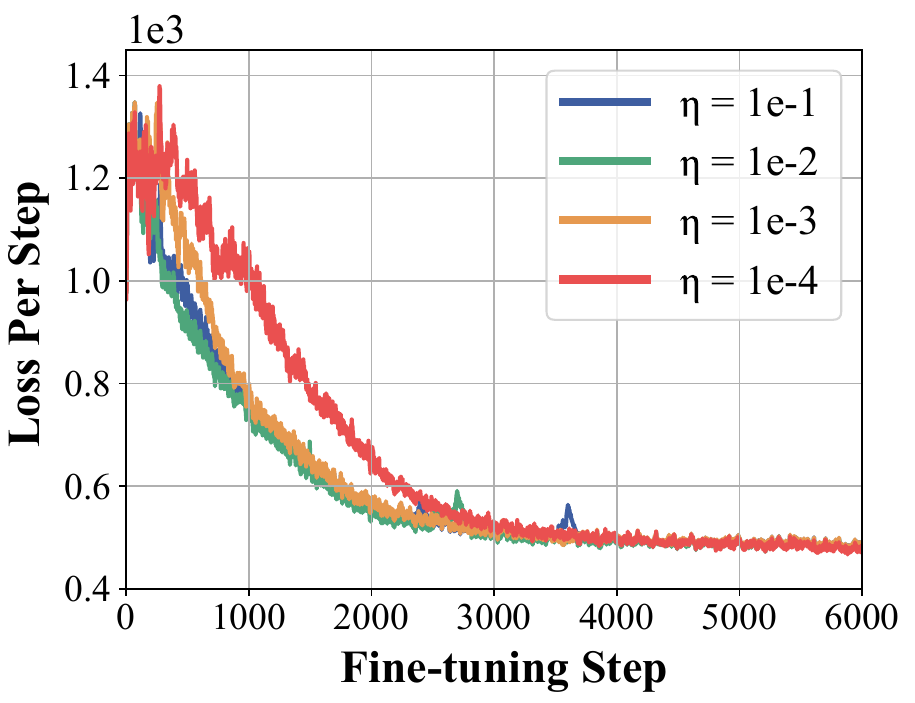}	
			\vspace{-6mm}
			\captionsetup{font=footnotesize}
			\caption*{(b) Learning Rate {\small $\eta$}}
			\label{fig:lr}
		\end{minipage}
	\end{tabular}
	\vspace{-2mm}
	\captionsetup{font=normalsize}
	\caption{Learning performance of \texttt{Tanbr} with T5-based MoE under different settings of parameters.
	}
	\label{fig:parameter}
	\vspace{-3mm}
\end{figure}

\textbf{\textit{Ablation Study}.} We analyze the sensitivity of our \texttt{Tanbr} to key hyperparameters.
Fig.~\ref{fig:parameter} presents the learning performance of the T5-based MoE under different settings of the smoothness parameter $\rho$ and the neural network learning rate {\small $\eta$}.
A smaller {\small $\rho$} makes \texttt{Tanbr} more conservative when partitioning the cover tree, requiring more observations before a node is expanded.
As seen in Fig. \ref{fig:parameter}, a smaller {\small $\rho$} shortens the initial random phase, since the merging weights exhibit less variation and allow faster fine-tuning. However, this leads to relatively lower final performance, as the decision space is insufficiently explored. In contrast, a larger {\small $\rho$} allows for better exploration of the decision space but results in longer initial random times and slower fine-tuning. Therefore, choosing an appropriate value of {\small $\rho$} is critical for balancing exploration and exploitation.
The learning rate {\small $\eta$} controls the magnitude of updates during the training of the neural network. A higher learning rate accelerates convergence but may cause instability, while a lower learning rate results in more stable training but slower progress. Tuning $\eta$ is essential to achieve both stability and efficient convergence.

\section{Conclusion}\label{sec:conclu}

In this work, we propose \texttt{Tanbr}, a tree-structured adaptive neural bandit router, to improve the efficiency of online inference with MoEs.
By utilizing task distributions, which are easily obtained in online environments,
\texttt{Tanbr} dynamically merges experts within a given pre-trained MoE in a task-aware manner, enabling a single merged expert to handle multiple tasks.
This is achieved through a tree-structured partitioning method that efficiently explores the continuous decision space, followed by a neural bandit approach that models the non-linear mapping between merging weights and performance rewards, and selects the optimal merging weight. Through both theoretical analysis and extensive experiments, we show that \texttt{Tanbr} achieves sublinear regret and significantly reduces memory usage and inference latency while maintaining a high accuracy.
These advantages make \texttt{Tanbr} particularly suitable for deploying MoE models in resource-constrained settings, such as edge networks.

To reduce interference among multiple experts \cite{he2023merging, li2024merge}, future work could explore grouping similar experts for merging. This can be achieved by applying advanced clustering techniques \cite{atalar2025neural} to dynamically form expert groups, which could improve both efficiency and accuracy across various applications.

\section*{Appendix} \label{sec:append}
\setlength{\columnsep}{0.12in}

\subsection{Proof of Lemma 1}
Besides the assumptions in Sec. \ref{sec:algo_analysis}, we also have the following assumption.

\begin{Assumption}\textnormal{\textbf{(Near-Optimality dimension)}}
    Let {\small $\epsilon = 3\nu_1 \rho^h$} and {\small $\epsilon' = \nu_2 \rho^h < \epsilon$}. For any subset of {\small $\epsilon$}-optimal nodes {\small $\mathcal{X}_{\epsilon} = \{ \boldsymbol{x} \in \mathcal{X} : f^* - f(\boldsymbol{x}) \leq \epsilon \}$}, there exists a constant {\small $C$} such that {\small $\mathcal{N}(\mathcal{X}_{\epsilon}, \ell, \epsilon') \leq C (\epsilon')^d$}, where {\small $d$} is the near-optimality dimension of {\small $f$} and {\small $\mathcal{N}(\mathcal{X}_{\epsilon}, \ell, \epsilon')$} is the {\small $\epsilon'$}-cover number of {\small $\mathcal{X}_{\epsilon}$} with respect to the dissimilarity measure {\small $\ell$}.
\end{Assumption}

Then, we characterize the complexity of the problem using the near-optimality dimension. We first establish an upper bound on the maximum depth of the partition tree in \texttt{Tanbr}.

\begin{Lemma}\label{lem_depth}
    Given the threshold {\small $\tau_h(t)$} defined in Eq. (\ref{tau}), the depth {\small $H(t)$} of the tree {\small $\mathcal{H}_t$} is bounded by:
    \vspace{-2mm}
    {\small
    \begin{equation}
        H(t) \leq H_{max}(T) \leq \frac{1}{1-\rho}\log\frac{T\nu_1^2}{C^2 \log(T/\delta)}.
    \end{equation}}
\vspace{-5mm}
\end{Lemma}

\noindent
\textbf{\textit{Proof}.}
Recall that a node {\small $(h, i)$} is expanded only when its pull count {\small $P_{h,i}(t)$} exceeds the threshold {\small $\tau_h(t)$}. To find the maximum possible depth {\small $H(t)$} of any node that may be expanded up to round {\small $t$}, we consider the condition under which a node can be expanded: {\small $\tau_h(t) \leq t$}. By substituting the expression for {\small $\tau_h(t)$} from Eq.~\eqref{tau}, we obtain:
{\small $
\frac{C^2 \log(t/\delta)}{\nu_1^2} \rho^{-2h} \leq t
\Rightarrow \rho^{-2h} \leq \frac{t \nu_1^2}{C^2 \log(t/\delta)}.
$}

Taking the logarithm on both sides, we have the depth {\small $H(t)$} bounded by:
{\small $
H(t) \leq \frac{1}{2 \log(1/\rho)} \log\left( \frac{t \nu_1^2}{C^2 \log(t/\delta)} \right).
$}
To simplify the upper bound to depend solely on {\small $T$}, we use the inequalities {\small $t \leq T$} and {\small $\log(t/\delta) \geq \log(T/\delta)$}. This gives:

{\small 
	\vspace{-3mm}
	\begin{align*}
	H(T) \leq & \frac{1}{2 \log(1/\rho)} \log( \frac{T \nu_1^2}{C^2 \log(T/\delta)}) 
	 \leq \frac{1}{1 - \rho} \log( \frac{T \nu_1^2}{C^2 \log(T/\delta)} ), \nonumber
\end{align*}
\vspace{-3mm}
}

\noindent
where the second inequality follows  {\small $2 \log(1/\rho) \geq 1 - \rho, \forall \rho \in (0, 1)$}. This completes the proof of Lemma \ref{lem_depth}.

Lemma \ref{lem_depth} ensures that the maximum depth of the tree does not exceed {\small $O(\log T)$}.
Furthermore, we derive an upper bound on the number of arms at any time slot, denoted by {\small $N$} as:

{\small 
\vspace{-2mm}
\begin{equation}
N_t \leq \frac{2^{1/(1-\rho)} t \nu_1^2}{C^2 \log(T/\delta)} \leq \frac{2^{1/(1-\rho)} T \nu_1^2}{C^2 \log(T/\delta)} = N.
\end{equation}
\vspace{-5mm}}

\subsection{Proof of Theorem 1}
First, we show the definition of the Neural Tangent Kernel (NTK) matrix and its effective dimension.

\begin{Definition}
    \label{def_1}
    Let {\small $\boldsymbol{M}$} denote the NTK matrix associated with {\small $\mathcal{X}_T$}. The effective dimension {\small $\tilde{d}$} of {\small $\boldsymbol{M}$}, with regularization parameter {\small $\lambda$}, is defined by: {\small $\tilde{d} = \frac{\log \det (\boldsymbol{I} + \boldsymbol{M} / \lambda)}{\log (1 + TN / \lambda)}.$}
\end{Definition}

The NTK matrix {\small $\boldsymbol{M}$} is constructed recursively, layer by layer, from the input to the output of the neural network~\cite{jacot2018neural}. The effective dimension {\small $\tilde{d}$} captures the intrinsic dimensionality of the input in the Reproducing Kernel Hilbert Space (RKHS) induced by the NTK. For additional details, we refer the reader to~\cite{pmlrzhou20a}.

\begin{Lemma}
\label{lemma_theta}
\textnormal{(Lemma 5.2 \cite{pmlrzhou20a})}
There exist positive constants {\small $\bar{C}_1, \bar{C}_2$} such that for any {\small $\delta \in (0,1)$}, if  {\small $\eta \leq \bar{C}_1 (TwL + w\lambda)^{-1}$} and  
{\small $
w \geq \bar{C}_2 \max \left\{ T^7 \lambda^{-7} L^{21} (\log w)^3, \lambda^{-\frac{1}{2}} L^{-\frac{3}{2}} (\log(TNL^2 / \delta))^{\frac{3}{2}} \right\}
$},
then with probability at least {\small $1 - \delta$}, we have: {\small $\|\boldsymbol{\xi}_t - \boldsymbol{\xi}_0\|_2 \leq 2 \sqrt{t / (w \lambda)}, \|\boldsymbol{\xi}^* - \boldsymbol{\xi}_t\|_{\boldsymbol{Z}_t} \leq \gamma_t / \sqrt{w}, \forall t \in \mathcal{T}.$}
\end{Lemma}

\begin{Lemma}
\label{lemma_f*(x)}
    Let {\small $(h_t^*,i_t^*)$} be the leaf node that contains $\boldsymbol{x}_t^*$.
    There exist positive constants {\small $\overline{C}_1, \overline{C}_2$} such that for any {\small $\delta \in (0, 1)$}, if {\small $\eta$} and {\small $w$} satisfy the same conditions as in Lemma \ref{lemma_theta}, then with probability at least {\small $1 - \delta$}, we have:
    \begin{align}
        \small
        f_1^*(\boldsymbol{x}_t^*) - f_1^*(\boldsymbol{x}_{h_t,i_t}) \leq  & 2\min \{ \gamma_{t-1}\| \boldsymbol{g}(\boldsymbol{x}_{h_t,i_t},\boldsymbol{\xi}_{t-1})/\sqrt{w}\|_{\boldsymbol{Z}_{t-1}}, \nonumber \\
        & 1\}  + 2\bar{\Gamma} + \nu_1\rho^{h_t},
    \end{align}
    where {\small $\bar{\Gamma} = \overline{C}_1 w^{-1/6} \sqrt{\log w} t^{2/3} \lambda^{-2/3} L^3  +\overline{C}_2(\sqrt{2\boldsymbol{m}^\top\boldsymbol{M}^{-1}\boldsymbol{m}}+2\sqrt{t/\lambda}) w^{-1/6} \sqrt{\log w} t^{1/6} \lambda^{-1/6} L^{7/2}$}.
\end{Lemma}

We now prove that the cumulative regret satisfies:
{\small
\begin{align}
    R_T = &\sum_{t \in \mathcal{T}} \left[f_1^*(\boldsymbol{x}_t^*) - f_1^*(\boldsymbol{x}_{h_t,i_t})\right] \\
    \leq & 2\sqrt{T \sum_{t \in \mathcal{T}} \min \{ \gamma_{t-1}^2\| \boldsymbol{g}(\boldsymbol{x}_{h_t,i_t},\boldsymbol{\xi}_{t-1})/\sqrt{w}\|_{\boldsymbol{Z}_{t-1}}^2,1\}}  \nonumber \\
    &+ 2T\bar{\Gamma} + \sum_{t \in \mathcal{T}}\nu_1\rho^{h_t} \nonumber \\
    \leq & 3\sqrt{T}\sqrt{\tilde{d} \log(1 + T N / \lambda)+2} \cdot  \nonumber \\
    & \left[\upsilon\sqrt{\tilde{d} \log(1 + T N / \lambda)+2-2\log \delta} \right. \nonumber \\
    & \left. +(\lambda+C_3TL)(1-\eta w\lambda)^{J/2}\sqrt{T/\lambda}+2\sqrt{\lambda S} \right]+T+1. \nonumber
\end{align}
\vspace{-3mm}
}

\noindent
where the first inequality follows from Lemma~\ref{lemma_f*(x)} and the Cauchy–Schwarz inequality; the second inequality follows by applying Lemma 5.4 of \cite{pmlrzhou20a}, valid for sufficiently large {\small $w$}. This completes the proof of Theorem~\ref{thm:regret}.

\subsection{Proof of Lemma 4}

For the difference between the estimated value and the real value, we have:

\vspace{-4mm}
{\small 
\begin{align}
\label{lemma_f(x)_1}
		&\| f(\boldsymbol{x},\boldsymbol{\xi}_{t-1}) - <\boldsymbol{g}(\boldsymbol{x},\boldsymbol{\xi}_0),\boldsymbol{\xi}^* - \boldsymbol{\xi}_0>\| \\ \nonumber
		= & \| f(\boldsymbol{x},\boldsymbol{\xi}_{t-1}) - f(\boldsymbol{x},\boldsymbol{\xi}_{0}) - <\boldsymbol{g}(\boldsymbol{x},\boldsymbol{\xi}_0),\boldsymbol{\xi}_{t-1} - \boldsymbol{\xi}_0> \\ \nonumber
		& -<\boldsymbol{g}(\boldsymbol{x},\boldsymbol{\xi}_0),\boldsymbol{\xi}^* - \boldsymbol{\xi}_{t-1}>\| \\ \nonumber
		\leq  & \overline{C}_1 w^{-1/6}\sqrt{\log w} t^{2/3} \lambda^{-2/3}L^3 + \| <\boldsymbol{g}(\boldsymbol{x},\boldsymbol{\xi}_0),\boldsymbol{\xi}^* - \boldsymbol{\xi}_{t-1}>\|,
	\end{align}}
\vspace{-4mm}

\noindent
where the first equality holds due to {\small $f(\boldsymbol{x},\boldsymbol{\xi}_{0})$} by the random initialization of {\small $\boldsymbol{\xi}_{0}$}, the inequality
holds due to Lemma 4.1 of~\cite{cao2019advances} with the fact {\small $\| \boldsymbol{\xi}_{t-1} - \boldsymbol{\xi}_{0}\|_2 \leq 2\sqrt{t/(w\lambda)}$}. For the second term, we have the following bound:

{\small 
\vspace{-4mm}
\begin{align}
\label{lemma_f(x)_2}
    &\| <\boldsymbol{g}(\boldsymbol{x},\boldsymbol{\xi}_0),\boldsymbol{\xi}^* - \boldsymbol{\xi}_{t-1}>\|  \\ 
    \leq & \| <\boldsymbol{g}(\boldsymbol{x},\boldsymbol{\xi}_{t-1}),\boldsymbol{\xi}^* - \boldsymbol{\xi}_{t-1}>\| + \nonumber \\ 
    & \| <\boldsymbol{g}(\boldsymbol{x},\boldsymbol{\xi}_0) - \boldsymbol{g}(\boldsymbol{x},\boldsymbol{\xi}_{t-1}),\boldsymbol{\xi}^* - \boldsymbol{\xi}_{t-1}>\| \nonumber \\
    \leq & \| \boldsymbol{\xi}^* - \boldsymbol{\xi}_{t-1}\|_{\boldsymbol{Z}_{t-1}} \| \boldsymbol{g}(\boldsymbol{x},\boldsymbol{\xi}_{t-1})\|_{\boldsymbol{Z}_{t-1}} +  \nonumber \\
    & \| \boldsymbol{\xi}^* - \boldsymbol{\xi}_{t-1}\|_2\|\boldsymbol{g}(\boldsymbol{x},\boldsymbol{\xi}_0) - \boldsymbol{g}(\boldsymbol{x},\boldsymbol{\xi}_{t-1}) \|_2  \nonumber \\ 
    \leq & \gamma_{t-1}\| \boldsymbol{g}(\boldsymbol{x},\boldsymbol{\xi}_{t-1})/\sqrt{w}\|_{\boldsymbol{Z}_{t-1}} + \nonumber \\
    &\| (\boldsymbol{\xi}^* - \boldsymbol{\xi}_{0}) + ( \boldsymbol{\xi}_{0} - \boldsymbol{\xi}_{t-1})\|_2\|\boldsymbol{g}(\boldsymbol{x},\boldsymbol{\xi}_0) - \boldsymbol{g}(\boldsymbol{x},\boldsymbol{\xi}_{t-1}) \|_2  \nonumber  \\  
    \leq & \gamma_{t-1}\| \boldsymbol{g}(\boldsymbol{x},\boldsymbol{\xi}_{t-1})/\sqrt{w}\|_{\boldsymbol{Z}_{t-1}} + \nonumber \\
    & \overline{C}_2(\sqrt{2\boldsymbol{m}^\top\boldsymbol{M}^{-1}\boldsymbol{m}}+2\sqrt{t/\lambda})w^{-1/6}\sqrt{\log w} t^{1/6}\lambda^{-1/6}L^{7/2}, \nonumber
\end{align}
\vspace{-4mm}}

\noindent
where the first inequality is derived from the triangle inequality; the second inequality holds due to Hölder’s inequality; the third inequality follows Lemma \ref{lemma_theta}; the fourth inequality uses Lemmas B.5 and B.6 in \cite{pmlrzhou20a}.
Combining Eq. (\ref{lemma_f(x)_1}) and Eq. (\ref{lemma_f(x)_2}), we obtain:

\vspace{-3mm}
\begin{align}
    \label{eq:f(x)}
    f(\boldsymbol{x},\boldsymbol{\xi}_{t-1}) - \langle \boldsymbol{g}(\boldsymbol{x},\boldsymbol{\xi}_0), \boldsymbol{\xi}^* - \boldsymbol{\xi}_0 \rangle \|
        \leq  \nonumber \\  \gamma_{t-1} \| \boldsymbol{g}(\boldsymbol{x},\boldsymbol{\xi}_{t-1})/\sqrt{w} \|_{\boldsymbol{Z}_{t-1}} +    \bar{\Gamma}.
\end{align}
\vspace{-4mm}

Now, we can compute:

{\small
	\vspace{-3mm}
\begin{align}
    & f_1^*(\boldsymbol{x}_t^*) - f_1^*(\boldsymbol{x}_{h_t,i_t})\\ \nonumber
    = &f_1^*(\boldsymbol{x}_t^*) -f_1^*(\boldsymbol{x}_{h_t^*,i_t^*}) + f_1^*(\boldsymbol{x}_{h_t^*,i_t^*}) - f(\boldsymbol{x}_{h_t^*,i_t^*},\boldsymbol{\xi}_{t-1}) + \\ \nonumber
    & f(\boldsymbol{x}_{h_t^*,i_t^*},\boldsymbol{\xi}_{t-1}) - f(\boldsymbol{x}_{h_t,i_t},\boldsymbol{\xi}_{t-1}) + f(\boldsymbol{x}_{h_t,i_t},\boldsymbol{\xi}_{t-1}) - f_1^*(\boldsymbol{x}_{h_t,i_t}) \\ \nonumber
    \leq & \|f_1^*(\boldsymbol{x}_t^*) -f_1^*(\boldsymbol{x}_{h_t^*,i_t^*}) \|+ \|f_1^*(\boldsymbol{x}_{h_t^*,i_t^*}) - f(\boldsymbol{x}_{h_t^*,i_t^*},\boldsymbol{\xi}_{t-1})\| + \\ \nonumber
    f(&\boldsymbol{x}_{h_t^*,i_t^*},\boldsymbol{\xi}_{t-1}) - f(\boldsymbol{x}_{h_t,i_t},\boldsymbol{\xi}_{t-1}) + \| f(\boldsymbol{x}_{h_t,i_t},\boldsymbol{\xi}_{t-1}) - f_1^*(\boldsymbol{x}_{h_t,i_t})\| \nonumber \\ 
    \leq & 2\gamma_{t-1}\| \boldsymbol{g}(\boldsymbol{x}_{h_t,i_t},\boldsymbol{\xi}_{t-1})/\sqrt{w}\|_{\boldsymbol{Z}_{t-1}}  + 2\bar{\Gamma}   + \nu_1\rho^{h_t} \nonumber \\
    \leq & \min\{ 2\gamma_{t-1}\| \boldsymbol{g}(\boldsymbol{x}_{h_t,i_t},\boldsymbol{\xi}_{t-1})/\sqrt{w}\|_{\boldsymbol{Z}_{t-1}}  + 2\bar{\Gamma}  + \nu_1\rho^{h_t}, 1\} \nonumber \\
    \leq & 2\min \{ \gamma_{t-1}\| \boldsymbol{g}(\boldsymbol{x}_{h_t,i_t},\boldsymbol{\xi}_{t-1})/\sqrt{w}\|_{\boldsymbol{Z}_{t-1}},1\}  + 2\bar{\Gamma} + \nu_1\rho^{h_t}, \nonumber
\end{align}
\vspace{-4mm}
}

\noindent
where the first inequality follows from the triangle inequality and the fact that the difference between any two scalar values can be upper bounded by the sum of their pairwise differences; the second inequality applies the local smoothness assumption and the bound on the prediction error, as given in  Eq. (\ref{eq:f(x)}); the third inequality holds because the reward function is bounded within {\small $[0, 1]$}, which can be achieved through normalization; the fourth inequality follows from the property that {\small $\min\{a + b, 1\} \leq \min\{a, 1\} + b$} for any non-negative values {\small $a$} and {\small $b$}.

\newpage
\bibliographystyle{IEEEtran}
\bibliography{IEEEabrv,reference}

\newpage

\end{document}